\PassOptionsToPackage{hyphens}{url}
\documentclass[final,onefignum,onetabnum]{siamonline171218}
\usepackage{amsfonts}
\usepackage{xfrac}
\usepackage{bm}
\usepackage{graphicx}
\usepackage{epstopdf}
\usepackage{algorithmic}
\usepackage{amssymb}
\usepackage{booktabs}
\usepackage{algorithmic}
\Crefname{ALC@unique}{Line}{Lines}
\usepackage[T1]{fontenc}
\usepackage{tikz}
\usepackage[caption=false,font=footnotesize]{subfig}
\usepackage{pgfplots}
\usepackage{tikz-network}
\pgfplotsset{compat=newest}
\pgfplotsset{plot coordinates/math parser=false}
\usepackage{url}
\usepackage{color}
\usepackage{adjustbox}
\usepackage{enumitem}
\usepackage{placeins} 

\usetikzlibrary{plotmarks}

\newlength\figureheight
\newlength\figurewidth 

\ifpdf
  \DeclareGraphicsExtensions{.eps,.pdf,.png,.jpg}
\else
  \DeclareGraphicsExtensions{.eps}
\fi

\def\norm#1{\|#1\|} 

\definecolor{darkgreen}{rgb}{0.0,0.5,0.0}
\definecolor{amber}{rgb}{1.0, 0.75, 0.0}

\def\norm#1{\left\|#1\right\|} 
\numberwithin{theorem}{section}

\newcommand{\TheTitle}{Semi-supervised Learning for Aggregated Multilayer Graphs Using Diffuse Interface Methods and Fast Matrix Vector Products}
\newcommand{\ShortTitle}{SSL for Aggregated Multilayer Graphs}
\newcommand{\TheAuthors}{Kai Bergermann, Martin Stoll, Toni Volkmer}

\headers{\ShortTitle}{\TheAuthors}

\title{{\TheTitle} 
}

\author{Kai Bergermann\thanks{Technische Universit\"at Chemnitz, Department of Mathematics, Chair of Numerics of Partial Differential Equations, 09107 Chemnitz, Germany, ({\tt kai.bergermann@mathematik.tu-chemnitz.de})}\and Martin Stoll\thanks{Technische Universit\"at Chemnitz, Department of Mathematics, Chair of Scientific Computing, 09107 Chemnitz, Germany, ({\tt martin.stoll@mathematik.tu-chemnitz.de})} \and Toni Volkmer\thanks{Technische Universit\"at Chemnitz, Department of Mathematics, Chair of Scientific Computing, 09107 Chemnitz, Germany, ({\tt toni.volkmer@mathematik.tu-chemnitz.de})}
}

\usepackage{amsopn}
\DeclareMathOperator{\diag}{diag}
\ifpdf
\hypersetup{pdftitle={\TheTitle},
  pdfauthor={\TheAuthors}
}
\fi

\newcommand{\p}{\mathcal{P}}

\def\norm#1{\left\|#1\right\|} 

\begin{document}

\maketitle

\begin{abstract}
 We generalize a graph-based multiclass semi-supervised classification technique based on diffuse interface methods to multilayer graphs. Besides the treatment of various applications with an inherent multilayer structure, we present a very flexible approach that interprets high-dimensional data in a low-dimensional multilayer graph representation. Highly efficient numerical methods involving the spectral decomposition of the corresponding differential graph operators as well as fast matrix-vector products based on the nonequispaced fast Fourier transform (NFFT) enable the rapid treatment of large and high-dimensional data sets.  We perform various numerical tests putting a special focus on image segmentation. In particular, we test the performance of our method on data sets with up to 10 million nodes per layer as well as up to 104 dimensions resulting in graphs with up to $52$ layers. While all presented numerical experiments can be run on an average laptop computer, the linear dependence per iteration step of the runtime on the network size in all stages of our algorithm makes it scalable to even larger and higher-dimensional problems.
\end{abstract}


\pagestyle{myheadings}
\thispagestyle{plain}
\markboth{}{}

\section{Introduction}

Complex networks have become an indispensable tool in the modeling of
phenomena ranging from neurobiology to statistical physics \cite{Strogatz2001Exploring}. As most sets of items interact in a variety of relationships, multilayer graphs have emerged as a flexible tool to reflect these complex interactions, see also \cite{Kivela2014MultilayerNetworks,boccaletti2014structure}. The power of these networks to model phenomena from social interactions to energy networks has greatly fueled research for a better understanding of the network properties and also to tailor numerical methods to incorporate their mathematical structures. 

In this paper, we propose a technique for semi-supervised learning, cf.\ \cite{zhou2004learning,zhu2009introduction} for an introduction, on multilayer graphs. Graph-based semi-supervised learning has recently risen to prominence with the introduction of graph convolutional neural networks \cite{Kipf:2016tc}. In other semi-supervised settings, the graph Laplacian enters via a regularization term \cite{hein2013total}. Besides that, the unsupervised case was also considered, e.g., the detection of communities without any labeled data in multiplex networks based on Laplacian dynamics was discussed in \cite{jeub2017local,Mucha2010community}. Furthermore, \cite{luxburg} serves as a survey about spectral clustering.

Our proposed method represents significant progress over existing methods that rely on spectral information of fully connected networks where the adjacency matrices are formed by evaluating kernel functions on the feature vectors representing the graph nodes. This is achieved by the incorporation of fast matrix-vector products with the graph Laplacian. In this paper, we discuss the extension of a recently introduced acceleration technique \cite{NFFTmeetsKrylov2018} to multilayer graphs. We find that the matrix-vector products with the graph Laplacian scale linearly not only in the number of graph nodes, but also in the feature space dimension when employing a feature grouping approach introduced later in this paper.

In particular we focus on semi-supervised learning where only around $0.5\%$ to $5\%$ of the data is pre-labeled.
In order to classify the remaining unlabeled graph nodes, we rely on a diffuse interface approach which was first introduced in \cite{BertozziFlenner2012DiffuseHighdimClassif}. This method by Bertozzi and Flenner borrows from well-known results that have mainly been studied in the context of phase separation phenomena in materials science \cite{onuki2002phase,wheeler1992phase,wodo2012modeling,bergermann2019modeling}. The crucial formulation on a graph then requires the use of a discrete differential operator, namely the graph  Laplacian \cite{luxburg,ChungSpectralGraphTheory1997}. Based on its properties and additional terms in the loss function, this method has shown great potential for different applications. This technique has recently been extended to various different scenarios including multiclass segmentation \cite{Garcia2014MulticlassDataSegm}, the use of an MBO scheme \cite{LuoBertozzi2017ConvergencyGraphAllenCahn,van2014mean} and of non-smooth potentials \cite{BoschKlamtStoll2018DiffuseInterface}, application to signed networks \cite{mercado2019node} and many more, see also~\cite{bertozzi2016diffuse} for an overview. Its extension and efficient implementation for the multilayer case is at the heart of this paper. 

Of course, we will require a corresponding differential operator, and we rely on the formulation of graph Laplacians for multilayer graphs. There exist many different representations for multilayer networks as well as different aggregation approaches \cite{Kivela2014MultilayerNetworks}. The use of an aggregated Laplacian that effectively combines the crucial information of the single layer graphs will be essential for our method, especially when considering artificially created multilayer structures using the feature grouping approach proposed in \Cref{sec:feature_grouping}. Encouraged by the results presented in~\cite{pmlr-v84-mercado18a}, we focus on the power mean Laplacian in this work. The use of this differential graph operator for semi-supervised learning was introduced in \cite{mercado2019generalized}, and we compare the numerical results of our scheme to those of that method.
For general data with an inherent multilayer structure, different tools like, e.g., supra-Laplacians with interlayer edges, cf.\ \cite[Section~2.3]{Kivela2014MultilayerNetworks} for the definition, can, from a numerical perspective, be used in combination with the methods introduced in this paper. Although methods for a separate treatment of inter- and intralayer edges might be required from an application point of view, this approach would in principle allow for labels to change over different layers, i.e., time in temporal multiplex networks \cite{Kivela2014MultilayerNetworks}. In this paper, however, we only consider classification problems where nodes are not allowed to change labels across different layers.

We demonstrate the performance of our method on synthetic as well as real data sets with a focus on the application of graph-based image segmentation, where the resulting graph is often fully connected and the sparsity of the network cannot be exploited. Additionally, the dimensionality of the feature space is often vast, making it a crucial task to be able to work with the resulting matrices in an efficient manner~\cite{stoll2020literature}. Many techniques in machine learning rely on the spectral information of the graph Laplacian in question, see also \cite{
calatroni2017graph,
KuSchmLoLeDLuAl10,
meng2017hyperspectral,
Mohar91thelaplacian,
Mohar1997}
in addition to the ones mentioned before. The diffuse interface method considered in this paper combines the favorable properties of the eigeninformation of the graph Laplacian with a nonlinear function pushing the graph nodes into their corresponding classes.
The computation of both eigenvalues and eigenvectors heavily relies on the efficiency of the matrix-vector products with the graph Laplacian. For moderate dimensions of the feature space, in particular $d\leq 3$,
and when the weight function of the graph is a smooth kernel function like a Gaussian kernel function,
fast summation techniques \cite{usingNFFT3,PoSt02,PoStNi04,NFFTmeetsKrylov2018,morariu2009automatic} show great potential to implicitly realize the matrix-vector multiplication in $\mathcal{O}(n)$ where $n$ is the number of nodes in the graph. These techniques are often based on arguments from Fourier analysis.

\begin{sloppypar}
In order to also take advantage of fast summation techniques for medium to high-dimensional data, e.g., ranging from 4 to more than 100 spatial dimensions, we present a feature grouping approach that splits the feature space into several low-dimensional subspaces. We interpret each feature subspace as a layer in a multilayer graph giving rise to a novel class of artificially created multilayer graphs which are then recombined again using our aggregated graph Laplacian. Each feature subspace can then take advantage of the fast matrix-vector products described before. This highly scalable technique not only enables us to efficiently classify large data sets like large images with several megapixels that would normally produce enormous graph Laplacian matrices but also allows for a high feature space dimensionality, which can be found in various applications, including for example hyperspectral imaging. 
\end{sloppypar}

We achieve the outlined tasks in this paper as follows. First, we give the necessary definitions for both graphs and multilayer graphs including the discrete differential operators in \Cref{sec:graphs_and_multilayer}, and we comment on eigenpairs computation approaches in \Cref{sec:eigenpairs_PKSM}. The fast summation technique based on the nonequispaced fast Fourier transform (NFFT), which is utilized in this paper, is introduced in \Cref{sec:nfft_fastsum}. We review the graph Allen--Cahn equation for (single layer) graphs in \Cref{sec:graph_allen_cahn_multiclass} focusing on the multiclass case. Its extension to the multilayer case is given in \Cref{sec:graph_allen_cahn_multilayer}. In \Cref{sec:feature_grouping} we propose the reformulation of a graph-based problem with a high-dimensional feature space as a multilayer graph via a feature grouping technique, allowing for the application of the NFFT-based fast summation introduced in \Cref{sec:nfft_fastsum}. Finally, 
the derived methods are applied to the classification of various data sets.
Several real world data sets with an inherent multilayer structure are considered in \Cref{sec:numerics:multilayer}, and large parts of the results obtained in \cite{mercado2019generalized} are further improved. \Cref{sec:numerics:image} then presents the segmentation of a 10 megapixel image, taking full advantage of fast matrix-vector products, and shows that our method generalizes very well to similar unseen images. 
Finally, in \Cref{sec:numerics:hyperspectral} all methods presented in this paper join forces in order to efficiently treat the both large and high-dimensional hyperspectral Pavia center data set \cite{plaza2009recent} with $148\,152$ pixels and 102 frequency bands achieving excellent classification accuracies while working directly on the unfiltered raw data without requiring problem-tailored hard- or software architectures. In particular, all numerical experiments can be run on a laptop computer.
Additionally in \Cref{sec:numerics:SBM}, some persuasive features of the power mean Laplacian are illustrated by classifying data sets generated by the stochastic block model \cite{holland1983stochastic}.

\section{Graphs and multilayer graphs}
\label{sec:graphs_and_multilayer}

First, we briefly introduce the notation of graphs and graph Laplacians. For more details and properties, we refer to \cite{Mohar91thelaplacian,Mohar1997,ChungSpectralGraphTheory1997}.

A graph $\mathcal{G}=(\mathcal{V},\mathcal{E})$ consists of vertices $x_i\in\mathcal{V}$, $|\mathcal{V}|=n$, and edges $e\in\mathcal{E}\subset\mathcal{V}\times\mathcal{V}$, where an edge $e$ connects any pair of vertices $x_{i}, x_{j} \in \mathcal{V}$. In this paper, we do not allow self-edges, i.e., we require $(x_i,x_i)\not\in\mathcal{E}$ $\forall i$. In particular, we use weighted graphs $\mathcal{G}$ with a weight function $w\colon\mathcal{V}\times\mathcal{V}\rightarrow\mathbb{R}_{\geq 0}$. A value $w(x_{i}, x_{j})>0$ indicates that two vertices $x_{i}, x_{j} \in \mathcal{V}$ are connected by an edge and $w(x_{i}, x_{j})=0$ means $(x_{i}, x_{j})\not\in\mathcal{E}$. The weight matrix $\boldsymbol{W}:=\left(w(x_i,x_j)\right)_{i,j=1}^n\in\mathbb{R}_{\geq 0}^{n\times n}$ collects the weight information. Here, we only consider undirected graphs, which yields $w(x_i,x_j)=w(x_j,x_i)$, leading to a symmetric weight matrix~$\boldsymbol{W}$. Since self-edges are not allowed, the diagonal of $\boldsymbol{W}$ only contains zeros.

Based on the weight function $w$, the degree $\operatorname{deg}(x_i)$ of a node $x_i$ can be defined as
\begin{equation*}
\operatorname{deg}(x_i) := \sum_{x_{j} \in \mathcal{V}} w(x_{i},x_{j})
\end{equation*}
and the degree matrix $\boldsymbol{D} \in \mathbb{R}^{n \times n}$ as the diagonal matrix 
\begin{equation*}
\boldsymbol{D}:=\diag\left(\operatorname{deg}(x_{1}),\ldots,\operatorname{deg}(x_{n})\right) = \diag\left(\boldsymbol{W}\boldsymbol{1}\right), \qquad \boldsymbol{1}:=(1,1,\ldots,1)^\top\in\mathbb{R}^n.
\end{equation*}
Then, the (unnormalized symmetric) graph Laplacian $\boldsymbol{L}\in\mathbb{R}^{n\times n}$ is defined as $\boldsymbol{L}:=\boldsymbol{D}-\boldsymbol{W}$ and the symmetric normalized graph Laplacian as
\begin{equation*}
 \boldsymbol{L}_{\mathrm{sym}} := \boldsymbol{D}^{-1/2} \boldsymbol{L} \boldsymbol{D}^{-1/2} = \boldsymbol{I} - \boldsymbol{D}^{-1/2} \boldsymbol{W} \boldsymbol{D}^{-1/2}.
\end{equation*}

The eigeninformation of these Laplacians plays a key role in many graph-based learning techniques, especially in classification tasks. For the efficient computation of this eigeninformation, repeated matrix-vector multiplications with the graph Laplacian, particularly with the weight matrix~$\boldsymbol{W}$, are required.

In order to be able to perform these tasks within a reasonable time frame for a large number~$n$ of nodes, the computation time for such matrix-vector products has to be reasonable. This can be achieved, in particular, if the weight matrix~$\boldsymbol{W}$ fulfills one of the following properties:
\begin{enumerate}[label=(\roman*)]
\item The weight matrix $\boldsymbol{W}\in\mathbb{R}_{\geq 0}^{n\times n}$ is dense and has no exploitable structure, leading to $\mathcal{O}(n^2)$ computation time for each matrix-vector multiplication with $\boldsymbol{W}$, but the number~$n$ of nodes is not too large. $\boldsymbol{W}$ can be either built once and stored, or it can be assembled on the fly for each matrix-vector multiplication.
\item The weight matrix $\boldsymbol{W}$ is sparse, e.g., containing only $\mathcal{O}(n)$ non-zero entries and resulting in $\mathcal{O}(n)$ computation time for a matrix-vector multiplication, when $\boldsymbol{W}$ is stored appropriately.
\item There exists a rank-$r$ factorization of $\boldsymbol{W}$ with low rank $r\ll n$, resulting in $\mathcal{O}(r\,n)$ computation time for each matrix-vector multiplication with $\boldsymbol{W}$.
\item A feature vector $\boldsymbol{x}_i\in\mathbb{R}^d$ is associated with each node $x_i\in\mathcal{V}$ of the graph~$\mathcal{G}$, and the weight function~$w$ is given by a suitable kernel function, $w(x_i,x_j) = K(\boldsymbol{x}_i-\boldsymbol{x}_j)$, such that the matrix-vector multiplication with~$\boldsymbol{W}$ can still be realized in $\mathcal{O}(n)$ computation time via a highly efficient algorithm without the need to explicitly store~$\boldsymbol{W}$, although~$\boldsymbol{W}$ may be densely populated. 
This case will be discussed in more detail in \Cref{sec:nfft_fastsum}.
\end{enumerate}

\begin{figure}
\begin{center}
\begin{tikzpicture}[multilayer=3d]
\SetLayerDistance{-2.2}
\Plane[x=2.3,y=6.5,width=4,height=3,layer=3]
\Plane[x=0.9,y=3,width=4,height=3,layer=2]
\Plane[x=-.5,y=-.5,width=4,height=3,layer=1]
\begin{Layer}[layer=1]
 \node at (.5,-.5)[below right]{Layer 1};
\end{Layer}
\begin{Layer}[layer=2]
 \node at (1.5,3)[below right]{Layer 2};
\end{Layer}
\begin{Layer}[layer=3]
 \node at (3.3,6.5)[below right]{Layer 3};
\end{Layer}

\Vertex[x=3.3,y=8.5,layer=3,IdAsLabel=false,label=$x_1$,size=.8,fontsize=\large]{31}
\Vertex[x=5.3,y=7.5,IdAsLabel=false,label=$x_2$,layer=3,size=.8,fontsize=\large]{32}
\Vertex[x=4.8,y=8.8,IdAsLabel=false,label=$x_3$,layer=3,size=.8,fontsize=\large]{33}
\Vertex[x=3.3,y=7.2,IdAsLabel=false,label=$x_4$,layer=3,size=.8,fontsize=\large]{34}
\Vertex[x=1.9,y=5,layer=2,IdAsLabel=false,label=$x_1$,size=.8,fontsize=\large]{21}
\Vertex[x=3.9,y=4,IdAsLabel=false,label=$x_2$,layer=2,size=.8,fontsize=\large]{22}
\Vertex[x=3.4,y=5.3,IdAsLabel=false,label=$x_3$,layer=2,size=.8,fontsize=\large]{23}
\Vertex[x=1.9,y=3.7,IdAsLabel=false,label=$x_4$,layer=2,size=.8,fontsize=\large]{24}
\Vertex[x=0.5,y=1.5,layer=1,IdAsLabel=false,label=$x_1$,size=.8,fontsize=\large]{11}
\Vertex[x=2.5,y=0.5,IdAsLabel=false,label=$x_2$,layer=1,size=.8,fontsize=\large]{12}
\Vertex[x=2,y=1.8,IdAsLabel=false,label=$x_3$,layer=1,size=.8,fontsize=\large]{13}
\Vertex[x=0.5,y=0.2,IdAsLabel=false,label=$x_4$,layer=1,size=0.8,fontsize=\large]{14}

\Edge[bend=-30,color=red](14)(12)
\Edge[bend=30,color=red](14)(11)
\Edge[bend=30,color=red](11)(13)
\Edge[bend=-15,color=red](11)(12)
\Edge[bend=30,color=red](21)(24)
\Edge[bend=30,color=red](22)(23)
\Edge[bend=30,color=red](31)(33)
\Edge[bend=15,color=red](33)(34)
\Edge[bend=30,color=red](32)(34)
\Edge[bend=-30,color=red](31)(34)
\end{tikzpicture}
\caption{Example of a multilayer network with $3$ layers and $4$ nodes}
 \label{fig:multilayernetwork}
\end{center}
\end{figure}
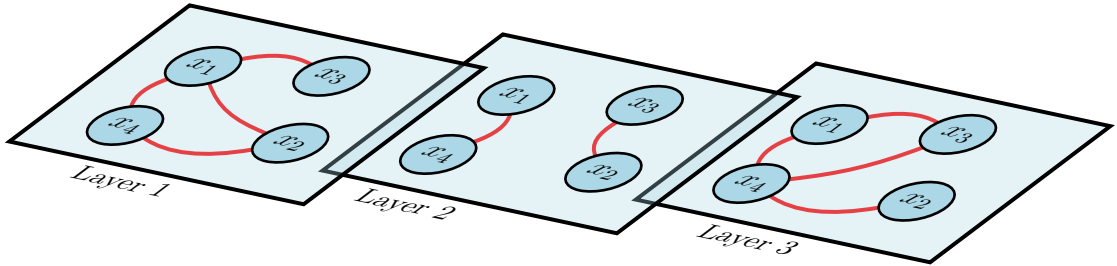

As many applications produce data graphs with an inherent multilayer structure, describing e.g., different types of interactions between nodes, time series data, or data sets combining data from independent sources \cite{Kivela2014MultilayerNetworks}, we next consider multilayer graphs, which consist of $T \in \mathbb{N}$ graph layers, see \Cref{fig:multilayernetwork} for an example. Now, each layer $\mathcal{G}^{(t)}, t=1, \dots , T$, is a graph based on the same vertex set $\mathcal{V}$, $|\mathcal{V}|=n$. The edge sets $\mathcal{E}^{(t)}\subset\mathcal{V}\times\mathcal{V}$, however, are typically different across the layers and correspondingly also the weight matrices $\boldsymbol{W}^{(t)} \in \mathbb{R}_{\geq 0}^{n \times n}$. In this paper, we do not allow interlayer edges.

Following~\cite{pmlr-v84-mercado18a}, the symmetric normalized graph Laplacian~$\boldsymbol{L}_{\mathrm{sym}}$ of a graph $\mathcal{G}$ can then be generalized to multilayer graphs.
$\boldsymbol{L}_{\mathrm{sym}}$ is now defined for each layer $t$ separately as $\boldsymbol{L}_{\mathrm{sym}}^{(t)}$. To merge the information of all graph layers into one Laplacian, the power mean Laplacian~$\boldsymbol{L}_{p}$, which is defined as
\begin{equation}\label{equ:power_mean_laplacian}
 \boldsymbol{L}_p := \left( \frac{1}{T} \sum_{t=1}^{T} (\boldsymbol{L}_{\mathrm{sym}}^{(t)})^p \right)^{1/p},
\end{equation}
was introduced in~\cite{pmlr-v84-mercado18a}, where $p\in\mathbb{R} \setminus \{ 0\}$ denotes matrix powers and is not meant elementwise. Similarly, for a positive definite matrix $\boldsymbol{A}$, the matrix $p$th root $\boldsymbol{A}^{1/p}$ is the unique positive definite solution of the matrix equation $\boldsymbol{X}^p=\boldsymbol{A}$, cf.~\cite{Higham2000FunctionsOfMatrices}. For $p>0$, the definition \eqref{equ:power_mean_laplacian} can be directly applied to the symmetric positive semi-definite graph Laplacians $\boldsymbol{L}_{\mathrm{sym}}^{(t)}$, where all eigenvalues are $\geq 0$ and at least one eigenvalue is zero, cf.\ \cite{luxburg} for this property. Note that for the $p=1$ case, similar constructions have already been used, see e.g.~\cite{tsuda2005fast,huang2012affinity}.

For $p<0$, however, \eqref{equ:power_mean_laplacian} conflicts with the non-invertibility of $\boldsymbol{L}_{\mathrm{sym}}^{(t)}$ due to the occurrence of at least one zero eigenvalue. To circumvent this issue, \cite[Section~2.2]{pmlr-v84-mercado18a} proposes to apply a diagonal shift of $\delta  \geq 0$ to each symmetric graph Laplacian to obtain a (strictly) positive definite version
\begin{equation*}
 \boldsymbol{L}_{\mathrm{sym},\delta}^{(t)} = \boldsymbol{L}_{\mathrm{sym}}^{(t)} + \delta \boldsymbol{I},
\end{equation*}
where the choice $\delta=\log (1+|p|)$ was suggested\footnote{While, from a theoretical perspective, an infinitesimal shift would suffice to guarantee the invertibility of $\boldsymbol{L}_{\mathrm{sym},\delta}^{(t)}$, the dependency on $p$ provides numerical stability as it prevents negative powers of the eigenvalues of $\boldsymbol{L}_{\mathrm{sym},\delta}^{(t)}$ to become large.} in~\cite{pmlr-v84-mercado18a} for $p<0$. For $p>0$ we set $\delta=0$. For the special case $p=0$, we refer to~\cite{pmlr-v84-mercado18a}. Combining the~$\boldsymbol{L}_{\mathrm{sym},\delta}^{(t)}$ for the different layers yields the shifted power mean Laplacian
\begin{equation} \label{equ:power_mean_laplacian_shifted}
 \boldsymbol{L}_{p,\delta} := \left( \frac{1}{T} \sum_{t=1}^{T} (\boldsymbol{L}_{\mathrm{sym},\delta}^{(t)})^p \right)^{1/p}.
\end{equation}

The benefit of considering negative powers $p$ in \eqref{equ:power_mean_laplacian_shifted} can be motivated by \cite{pmlr-v84-mercado18a}, where the convergence of $\boldsymbol{L}_{p,\delta}$ to the \texttt{AND}- and \texttt{OR}-operator on the presence of a common clustering structure across the layers is proven for $p \rightarrow \infty$ and $p \rightarrow - \infty$, respectively, in certain stochastic block model situations. With this interpretation, the power mean Laplacian tends to capture the underlying clustering structure even if it is only present in some graph layers for $p<0$, while the $p>0$ case requires a more uniform distribution of the clustering information across the layers. Thus, $\boldsymbol{L}_{p,\delta}$ should be more robust w.r.t.\ noisy layers for negative $p$. We illustrate this at an example in \Cref{sec:numerics:SBM}.

Note, that for real symmetric matrices $\boldsymbol{A}\in\mathbb{R}^{n \times n}$ with eigendecomposition $\boldsymbol{A}=\boldsymbol{\Phi} \boldsymbol{\Lambda} \boldsymbol{\Phi}^\top$, the result of a matrix function $f(\boldsymbol{A})$, defined as in \cite[Definition 1.1]{Higham2000FunctionsOfMatrices}, can be obtained via $f(\boldsymbol{A})=\boldsymbol{\Phi} f(\boldsymbol{\Lambda}) \boldsymbol{\Phi}^\top$, where $f(\boldsymbol{\Lambda})=\diag(f(\lambda_i)_{i=1}^n)$.
This means $f$ only acts on the eigenvalues.
In particular, $\lambda^p$ is an eigenvalue of the matrix power $\boldsymbol{A}^p$ if $\lambda$ is an eigenvalue of $\boldsymbol{A}$ and $1-\lambda$ is an eigenvalue of $\boldsymbol{I}-\boldsymbol{A}$, cf.\ e.g.~\cite{Higham2000FunctionsOfMatrices}.

\section{Computation of eigenpairs and Polynomial Krylov Subspace Method}
\label{sec:eigenpairs_PKSM}

It is well known that for many classification tasks involving graph Laplacians, the computation of its $k$ smallest eigenpairs is of particular interest as they provide optimal solutions to different graph cut problems, cf.\ e.g.~\cite{luxburg}.
To this end, when using the power mean Laplacian $\boldsymbol{L}_1$ or the shifted version $\boldsymbol{L}_{p,\delta}$, $p<0$, one needs to perform matrix-vector products with this matrix or the $p$th power
\begin{equation}\label{equ:L_p_delta_power_p}
 \boldsymbol{L}_{p,\delta}^p = \frac{1}{T} \sum_{t=1}^{T} (\boldsymbol{L}_{\mathrm{sym},\delta}^{(t)})^p.
\end{equation}

In the case $p=1$,
one has $\boldsymbol{L}_1 = \frac{1}{T} \sum_{t=1}^{T} \boldsymbol{L}_{\mathrm{sym}}^{(t)}$,
and the relevant eigeninformation of $\boldsymbol{L}_1$ can be obtained by computing the eigenpairs for the $k$ largest eigenvalues of
\begin{equation}
\label{equ:I-L_1}
\boldsymbol{I} - \boldsymbol{L}_1
=
\frac{1}{T} \sum_{t=1}^{T} (\boldsymbol{D}^{(t)})^{-1/2} \, \boldsymbol{W}^{(t)} \, (\boldsymbol{D}^{(t)})^{-1/2}
\end{equation}
via the Lanczos method \cite{golub2012matrix}. As mentioned in \Cref{sec:graphs_and_multilayer},
the resulting eigenvectors $\boldsymbol{\phi}$ are identical with the ones of $\boldsymbol{L}_1$ and the eigenvalues $\lambda$ of \eqref{equ:I-L_1} correspond to $1-\lambda$ of $\boldsymbol{L}_1$.

Likewise, in the case of general $p$ for a given eigenvalue $\lambda$ and eigenvector $\boldsymbol{\phi}$ of the real symmetric matrix~$\boldsymbol{L}_{p,\delta}$, the corresponding eigenvalue of $\boldsymbol{L}_{p,\delta}^p$ is $\lambda^p$ and the eigenvector remains~$\boldsymbol{\phi}$, cf.\ \Cref{sec:graphs_and_multilayer}. Since the function $f(\lambda)=\lambda^p$ is order reversing for $p<0$, it holds for $\lambda_{1} \leq \lambda_{2} \leq \dots \leq \lambda_{n}$ that $\lambda_1^p \geq \lambda_2^p \geq \dots \geq \lambda_n^p$. In order to obtain the first $k$ smallest eigenvalues $\lambda_{1}, \dots , \lambda_{k}$ and corresponding eigenvectors $\boldsymbol{\phi}_1,\ldots,\boldsymbol{\phi}_k$ in the case $p<0$, it is sufficient to compute the $k$ largest eigenvalues $\lambda_1^p, \dots , \lambda_k^p$ with its eigenvectors of~$\boldsymbol{L}_{p,\delta}^p$.
For this, we propose to utilize the Lanczos method, which requires matrix-vector multiplications of the graph Laplacian matrices~$(\boldsymbol{L}_{\mathrm{sym},\delta}^{(t)})^p$, cf.~\eqref{equ:L_p_delta_power_p}.
The latter can be realized by using the Polynomial Krylov Subspace Method (PKSM)~\cite{pmlr-v84-mercado18a}. For this we again rely on the Lanczos method for $\boldsymbol{L}^{(t)}_{\mathrm{sym},\delta}$, cf.\ \cite[Chapter~13]{Higham2000FunctionsOfMatrices}.

As with any Krylov subspace method, the main algorithmic cost comes from the matrix-vector products, i.e., multiplication with $\boldsymbol{L}^{(t)}_{\mathrm{sym},\delta}$. The matrix function is then approximated using a projected matrix of drastically reduced dimension for which well--established methods from dense linear algebra can be employed.

When the weight functions $w^{(t)}$ and, correspondingly, the weight matrices~$\boldsymbol{W}^{(t)}$ of the layers $t\in\{1,\ldots,T\}$ have a special structure, the matrix-vector multiplications and eigenpair computations can be accelerated considerably as we discuss in the next section.

\section{NFFT-based fast summation for fast matrix-vector multiplications with the weight matrix}
\label{sec:nfft_fastsum}

For a graph $\mathcal{G}=(\mathcal{V},\mathcal{E})$, the nodes $x_i\in\mathcal{V}$ are identified with feature vectors $\boldsymbol{x}_i\in\mathbb{R}^d$, and the weight function $w\colon\mathcal{V}\times\mathcal{V}\rightarrow\mathbb{R}_{\geq 0}$ is associated with a kernel function $K\colon\mathbb{R}^d\rightarrow\mathbb{R}_{\geq 0}$, $w(x_i,x_j)=K(\boldsymbol{x}_i-\boldsymbol{x}_j)$ for $\boldsymbol{x}_i\neq\boldsymbol{x}_j$. Since self-loops are excluded, we have $w(x_i,x_i)=0$ due to $(x_i,x_i)\not\in\mathcal{E}$. The matrix-vector multiplications with the weight matrix $\boldsymbol{W}\in\mathbb{R}_{\geq 0}^{n\times n}$ can then be sped up dramatically, especially if the matrix is non-sparse, e.g., when $K$ can be well approximated by a trigonometric polynomial achieving a desired tolerance.
For instance, $K$ could be the Gaussian radial basis function (RBF) kernel $K(\boldsymbol{y})=\exp (-\norm{\boldsymbol{y}}^2/\sigma^2 )$ or Laplacian RBF kernel $K(\boldsymbol{y})=\exp (-\norm{\boldsymbol{y}}/\sigma )$.
For dimensions $d\leq 3$, one very efficient method is the NFFT-based fast summation \cite{PoSt02,PoStNi04} which achieves a runtime complexity of $\mathcal{O}(n)$ for fixed accuracy. Subsequently, we briefly describe the general ideas and give a fast algorithm. For more details, we refer to \cite[Section~3]{NFFTmeetsKrylov2018}.
In case of higher dimensions $d\geq 4$\footnote{Here, the curse of dimensionality makes computations burdensome as the runtime depends exponentially on the feature space dimension as will be discussed later in this section. Note that the threshold $d \geq 4$ may shift to larger values of $d$ in the future when increased computing power becomes available.}, we propose a feature grouping approach later in \Cref{sec:feature_grouping}.

For technical reasons\footnote{On the one hand,
one has $w(x_i,x_i)=0$ since $(x_i,x_i)\not\in\mathcal{E}$. On the other hand, $K(\boldsymbol{x}_i-\boldsymbol{x}_i)=K(\boldsymbol{0})\neq 0$ in general. Modifying the kernel function (only) at the origin is not an option since $K$ should be smooth (or at least continuous).}, see also \cite[Section~3]{NFFTmeetsKrylov2018},
we consider the modified weight function $\tilde{w}(x_i,x_j)=K(\boldsymbol{x}_i-\boldsymbol{x}_j)$ $\forall x_i,x_j\in\mathcal{V}$ with associated matrix
\begin{equation*}
\tilde{\boldsymbol{W}} := \boldsymbol{W} + K(\boldsymbol{0})\, \boldsymbol{I},
\end{equation*}
and we have $\boldsymbol{W} = \tilde{\boldsymbol{W}} - K(\boldsymbol{0})\, \boldsymbol{I}$.
Then, the matrix-vector multiplication of~$\boldsymbol{W}$ with an arbitrary vector $\boldsymbol{v}\in\mathbb{C}^n$ can be written as
$
 \boldsymbol{W} \boldsymbol{v} = \tilde{\boldsymbol{W}} \boldsymbol{v} - K(\boldsymbol{0}) \, \boldsymbol{v}.
$ 
The last part $K(\boldsymbol{0}) \, \boldsymbol{v}$ is simply a multiplication of a scalar value with a vector,
and we will compute the first part $\tilde{\boldsymbol{W}} \boldsymbol{v}$ in a fast way using the NFFT-based fast summation.
Each entry of the result $\tilde{\boldsymbol{W}}\boldsymbol{v}$ reads as
\begin{equation}
\label{eq::nfftWtilde_x:kernel}
\left(\tilde{\boldsymbol{W}}\boldsymbol{v}\right)_i = f(\boldsymbol{x}_i) := \sum_{j=1}^n v_j \, K(\boldsymbol{x}_i-\boldsymbol{x}_j).
\end{equation}
The key idea for the efficient computation of~\eqref{eq::nfftWtilde_x:kernel} uses methods from Fourier analysis \cite{PlPoStTa18}. In particular, the kernel function $K$ is approximated by a $d$-variate trigonometric polynomial $K_\mathrm{RF}$, which allows to separate the computations involving the nodes~$x_i$ and~$x_j$ in~\eqref{eq::nfftWtilde_x:kernel:RF}.\footnote{We remark that in $K_\mathrm{RF}$, the subscript R stands for regularized kernel function (smooth 1-periodic) and F for the Fourier approximation by a trigonometric polynomial. The main trick -- from a linear algebra point of view -- is that one diagonalizes the (modified) weight matrix~$\tilde{\boldsymbol{W}}$ (in a certain sense) and computes $\tilde{\boldsymbol{W}}\boldsymbol{v}\approx \boldsymbol{A} \big(\operatorname{diag}(\hat{b}_{\boldsymbol{l}})_{\boldsymbol{l}\in I_N}\big) \boldsymbol{A}^* \boldsymbol{v}$ (cf.\ \eqref{eq::K_RF} for the definitions of $\hat{b}_{\boldsymbol{l}}$ and $I_N$). Here $\boldsymbol{A}:=(\mathrm{e}^{2\pi\mathrm{i}\boldsymbol{l} \boldsymbol{x_j}})_{j=1,\ldots,n;\;\boldsymbol{l}\in I_N}$ is the non-equispaced Fourier matrix, which can be applied to a vector using $\mathcal{O}((\texttt{m}_\text{NFFT})^{d} \, n + d\,2^d N^d \log N)$ arithmetic operations. Please note that, assuming one evaluation of $K$ has complexity $\mathcal{O}(d)$, one can evaluate $K$ at all $N^d$ nodes $\boldsymbol{k}/N$, $\boldsymbol{k}\in I_N$, and then compute by $d$-dimensional FFT the Fourier coefficients $\hat{b}_{\boldsymbol{l}} := N^{-d} \sum_{\boldsymbol{k}\in I_N} K(\boldsymbol{k}/N) \, \mathrm{e}^{-2\pi\mathrm{i}\boldsymbol{l}\boldsymbol{k}/N}$, $\boldsymbol{l}\in I_N$, in $\mathcal{O}(d\,N^d \log N)$ runtime complexity.} This will be one main ingredient for reducing the computational complexity from $\mathcal{O}(n^2)$ to $\mathcal{O}(n)$.
Assuming we have a suitable approximation
of $K$, e.g., up to a tolerance of $10^{-4}$ with respect to the $L^\infty$ norm\footnote{For the formulation of the tolerance the kernel function $K$ is assumed to be normalized, i.e.\ $|K(x)| \leq 1$. For more details, we refer to \cite{PoSt02} and \cite[Section~3]{NFFTmeetsKrylov2018}.}, given by
\begin{equation}\label{eq::K_RF}
 K(\boldsymbol{y})  \approx
 K_\mathrm{RF}(\boldsymbol{y}) := \sum_{\boldsymbol{l}\in I_N} \hat{b}_{\boldsymbol{l}} \,\mathrm{e}^{2\pi\mathrm{i}\boldsymbol{l} \boldsymbol{y}}, \quad
 I_N := \{-N/2,-N/2+1,\ldots,N/2-1\}^d,
\end{equation}
with bandwidth $N\in 2\mathbb{N}$ and Fourier coefficients~$\hat{b}_{\boldsymbol{l}}$,
we replace $K$ by $K_\mathrm{RF}$ in~\eqref{eq::nfftWtilde_x:kernel} and we obtain
\begin{align}
 \nonumber
 \left(\tilde{\boldsymbol{W}}\boldsymbol{x}\right)_i
 = f(\boldsymbol{x}_i)
 \approx f_\mathrm{RF}(\boldsymbol{x}_i)
 :=& \sum_{j=1}^{n}v_j \, K_\mathrm{RF}(\boldsymbol{x}_i-\boldsymbol{x}_j)
 = \sum_{j=1}^{n} v_j \sum_{\boldsymbol{l}\in I_N} \hat{b}_{\boldsymbol l} \,\mathrm{e}^{2\pi\mathrm{i}\boldsymbol{l} (\boldsymbol{x}_i-\boldsymbol{x}_j)} \\
 \label{eq::nfftWtilde_x:kernel:RF}
 =& \sum_{\boldsymbol{l}\in I_N} \hat{b}_{\boldsymbol l} \underbrace{\left(\sum_{j=1}^{n} v_j \,\mathrm{e}^{-2\pi\mathrm{i}\boldsymbol{l} \boldsymbol{x}_j}\right)}_{=:\hat{f}_{\boldsymbol{l}}} \,\mathrm{e}^{2\pi\mathrm{i}\boldsymbol{l} \boldsymbol{x}_i}, \quad\forall i=1,\ldots,n.
\end{align}
Comparing~\eqref{eq::nfftWtilde_x:kernel:RF} with the initial problem of evaluating~\eqref{eq::nfftWtilde_x:kernel}, the Fourier approximation~$K_\mathrm{RF}$ of the kernel function~$K$ has so far only introduced an additional sum over~$I_N$. In situations of large $n$ (starting at ten thousands ranging to millions and billions) however, the NFFT \cite{nfft3} manages to substantially speed up the evaluation of the inner sums $\hat{f}_{\boldsymbol{l}}:=\left( \sum_{j=1}^{n} v_{j} e^{-2 \pi \mathrm{i} \boldsymbol{l} \boldsymbol{x}_{j}} \right)$, $\boldsymbol{l}\in I_N$, as well as the computation of the outer sums $f_\mathrm{RF}(\boldsymbol{x}_i):=\sum_{\boldsymbol{l}\in I_N} \hat{b}_{\boldsymbol l}\,\hat{f}_{\boldsymbol l}$, $i=1,\ldots,n$.
We remark that one can not apply standard (equispaced) fast Fourier transform (FFT) in most cases since the feature vectors $\boldsymbol{x}_i$ are not located on an equispaced grid.

Therefore, the NFFT is the second main ingredient for the reduction in the computational complexity from $\mathcal{O}(n^2)$, when using the direct summation~\eqref{eq::nfftWtilde_x:kernel}, to $\mathcal{O}\big((\texttt{m}_\text{NFFT})^{d} \, n + d \, 2^d N^{d} \log N\big)$ for computing~\eqref{eq::nfftWtilde_x:kernel:RF} via the NFFT-based fast summation, where $\texttt{m}_\text{NFFT}$ represents an internal window cut-off parameter of the NFFT controlling the desired precision\footnote{In practice, $\texttt{m}_\text{NFFT}$ is independent of $n$ and typically chosen between 2 and 6. $\texttt{m}_\text{NFFT}=8$ yields errors close to machine precision for IEEE double precision, see also \cite[Section~5.2]{usingNFFT3}.}, cf.\ \cite[Section~3]{KuPo06}. In situations where the number $n$ of nodes of the graph is large and the feature space dimension $d$ is not too big, this represents a crucial gain in computational complexity.

The accuracy of the Fourier approximation~$K_\mathrm{RF}$ of the kernel function~$K$ depends on the decay of the Fourier coefficients of~$K$, which are influenced by smoothness properties of the kernel function. For instance, smooth rotational invariant kernel functions are particularly suited, e.g.\ the already mentioned Gaussian RBF kernel $K(\boldsymbol{y})=\exp (-\norm{\boldsymbol{y}}^2/\sigma^2 )$, Laplacian RBF kernel $K(\boldsymbol{y})=\exp (-\norm{\boldsymbol{y}}/\sigma )$ and many others. The Fourier coefficients~$\hat{b}_{\boldsymbol l}$ of the Fourier approximation~$K_\mathrm{RF}$ can be computed easily by sampling the kernel function~$K$ or a regularized version of~$K$ on an equispaced grid and applying an FFT which takes $\mathcal{O}(d\,N^{d} \log N)$ arithmetic operations, assuming that $K$ can be evaluated in $\mathcal{O}(d)$ arithmetic operations, see also \cite[Section~3]{NFFTmeetsKrylov2018} for more details.
Since the parameter~$\texttt{m}_\text{NFFT}$ and the bandwidth~$N$ only depend on the desired accuracy, we have a computational complexity of $\mathcal{O}(n)$ for fixed~$d$ and fixed accuracy.
For further technical details, we refer to \cite{PoSt02} and \cite[Section~3]{NFFTmeetsKrylov2018}.

\begin{sloppypar}
In total, we have a fast approximate algorithm for the matrix-vector multiplication $\tilde{\boldsymbol{W}}\boldsymbol{v}$ of complexity $\mathcal{O}(n)$ available, cf.\ \Cref{alg::nfft_fastsum}.
This algorithm is implemented as \texttt{applications/fastsum} and \texttt{matlab/fastsum} in C and \textsc{MATLAB} within the NFFT3 software library and freely available, see~\cite{nfft3}. Then, one easily computes $\boldsymbol{W} \boldsymbol{v}$ from $\tilde{\boldsymbol{W}}\boldsymbol{v}$ by subtracting the vector $K(\boldsymbol{0}) \, \boldsymbol{v}$.
\end{sloppypar}

\begin{algorithm}[htb!]
\caption{(\cite[Algorithm~1]{NFFTmeetsKrylov2018}). Fast approximate matrix-vector multiplication $\tilde{\boldsymbol{W}}\boldsymbol{x}$ using NFFT-based fast summation,
$(\tilde{\boldsymbol{W}}\boldsymbol{v})_i=\sum_{j=1}^{n}v_j \,K(\boldsymbol{x}_i-\boldsymbol{x}_j)$ $\forall i=1,\ldots,n$, \newline e.g.\
$K(\boldsymbol{x}_i-\boldsymbol{x}_j) = \exp (-\norm{\boldsymbol{x}_i-\boldsymbol{x}_j}^2/\sigma^2)$.}
\label{alg::nfft_fastsum}
\vspace{0.5em}
  \begin{tabular}{p{1.4cm}p{3.5cm}p{9.45cm}}
    Input:
                & $\left(\hat{b}_{\boldsymbol{l}}\right)_{\boldsymbol{l}\in I_N}$ & Fourier coefficients of trigonometric polynomial $K_\mathrm{RF}$ which approximates $K$, \\[0.4em]
                & $\left\lbrace \boldsymbol{x}_i\right\rbrace_{i=1}^{n}$ & nodes, $\boldsymbol{x}_i\in\mathbb{R}^d$, $\norm{\boldsymbol{x}_i} \in [-1/4, 1/4]^{d}$, \\[0.4em]
                & $\boldsymbol{v}=[v_1,v_2,\ldots,v_n]^T$  & vector $\in\mathbb{R}^n$. \\
  \end{tabular}
\vspace{1em}
  \begin{enumerate}
  \item Apply $d$-dimensional adjoint NFFT, see also \cite[Section~2.3]{usingNFFT3}, on $\boldsymbol{v}$ and obtain \newline $\hat{v}_{\boldsymbol{l}} \,\approx\, \sum_{j=1}^{n} v_j \,\mathrm{e}^{-2\pi\mathrm{i}\boldsymbol{l} \boldsymbol{x}_j}$ $\forall \boldsymbol{l}\in I_N$.
  \item Multiply result by Fourier coefficients $\left(\hat{b}_{\boldsymbol{l}}\right)_{\boldsymbol{l}\in I_N}$ of $K_\mathrm{RF}$ and obtain $\hat{f}_{\boldsymbol{l}} := \hat{b}_{\boldsymbol{l}} \, \hat{v}_{\boldsymbol{l}}$ $\forall \boldsymbol{l}\in I_N$.
  \item Apply $d$-dimensional NFFT on $\left(\hat{f}_{\boldsymbol l}\right)_{\boldsymbol{l}\in I_N}$ and obtain output \newline $\tilde{f}_\mathrm{RF}(\boldsymbol{x}_i) \,\approx\, \sum_{\boldsymbol{l}\in I_N} \hat{f}_{\boldsymbol{l}} \,\mathrm{e}^{2\pi\mathrm{i}\boldsymbol{l} \boldsymbol{x}_i}$ $\forall i=1,\ldots,n$.
  \end{enumerate}
\vspace{1em}
  \begin{tabular}{p{1.4cm}p{3.5cm}p{9.45cm}}
    Output: & $\Big\lbrack \tilde{f}_\mathrm{RF}(\boldsymbol{x}_i)\Big\rbrack_{i=1,\ldots,n}$ & $\tilde{f}_\mathrm{RF}(\boldsymbol{x}_i) \approx (\tilde{\boldsymbol{W}}\boldsymbol{v})_i$ $\forall i=1,\ldots,n$. \\
    \cmidrule{1-3}
 \end{tabular}
  \begin{tabular}{p{2.1cm}p{10.0cm}}
    Complexity: & $\mathcal{O}\big(n\big)$ for fixed accuracy. \\
  \end{tabular}
\end{algorithm}

Note that \Cref{alg::nfft_fastsum} can be used for accelerating the eigenpair computation of the power mean Laplacian~$\boldsymbol{L}_1$ and of the shifted power mean Laplacian~$\boldsymbol{L}_{p,\delta}^p$, $p<0$. In the latter case, this means applying \Cref{alg::nfft_fastsum} inside the Polynomial Krylov Subspace Method, see also \Cref{sec:eigenpairs_PKSM}.

In the next section, the eigeninformation of $\boldsymbol{L}_1$ and $\boldsymbol{L}_{p,\delta}^p$, $p<0$, is used in order to perform semi-supervised learning based on multilayer graphs.

\section{Graph Allen--Cahn for multiclass problems}\label{sec:graph_allen_cahn_multiclass}

\begin{figure}
\begin{center}
	\frame{\includegraphics*[width=0.485\textwidth]{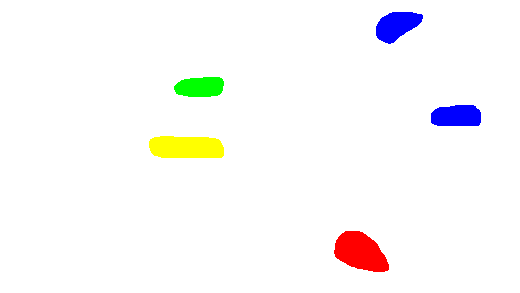}}
\hfill
	\frame{\includegraphics*[width=0.485\textwidth]{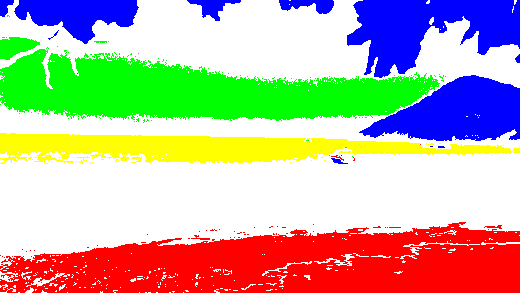}} \\[1em]
	\frame{\includegraphics*[width=0.485\textwidth]{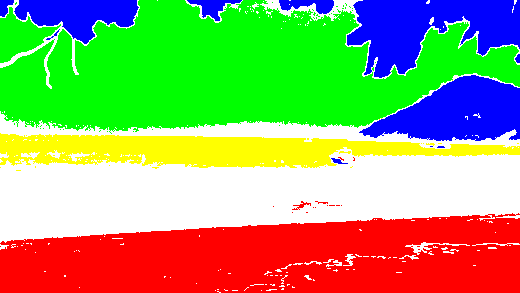}}
\hfill
	\frame{\includegraphics*[width=0.485\textwidth]{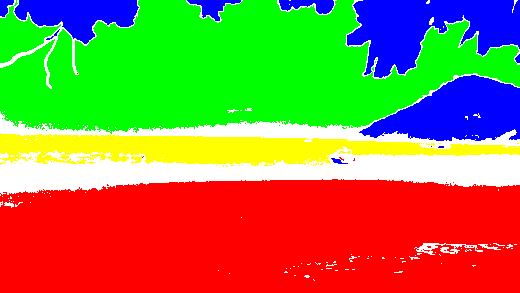}}
\end{center}
\caption{Example of the evolution of a phase-field simulation for image segmentation based on Graph Allen--Cahn with four classes, where the four colors red, green, blue, and yellow indicate different class affiliations with score values higher than 0.66. The image in this example is a resized version ($520 \times 293$ pixels) of \Cref{fig:beach_image}. \label{fig:beach_evolution}}
\end{figure}

\begin{figure}
\begin{center}
\includegraphics[width=.485\textwidth]{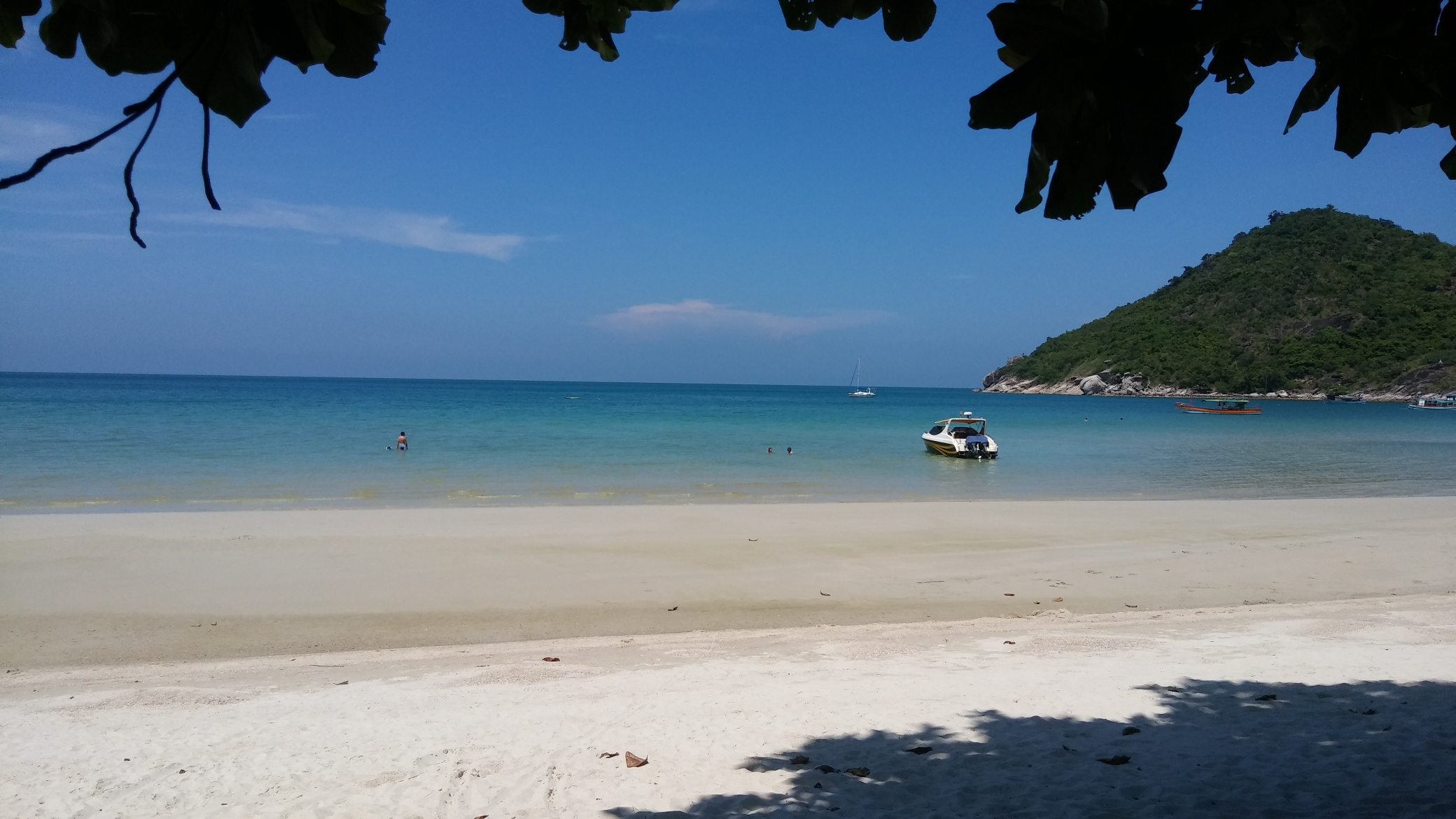}
\hfill
\includegraphics[width=.485\textwidth]{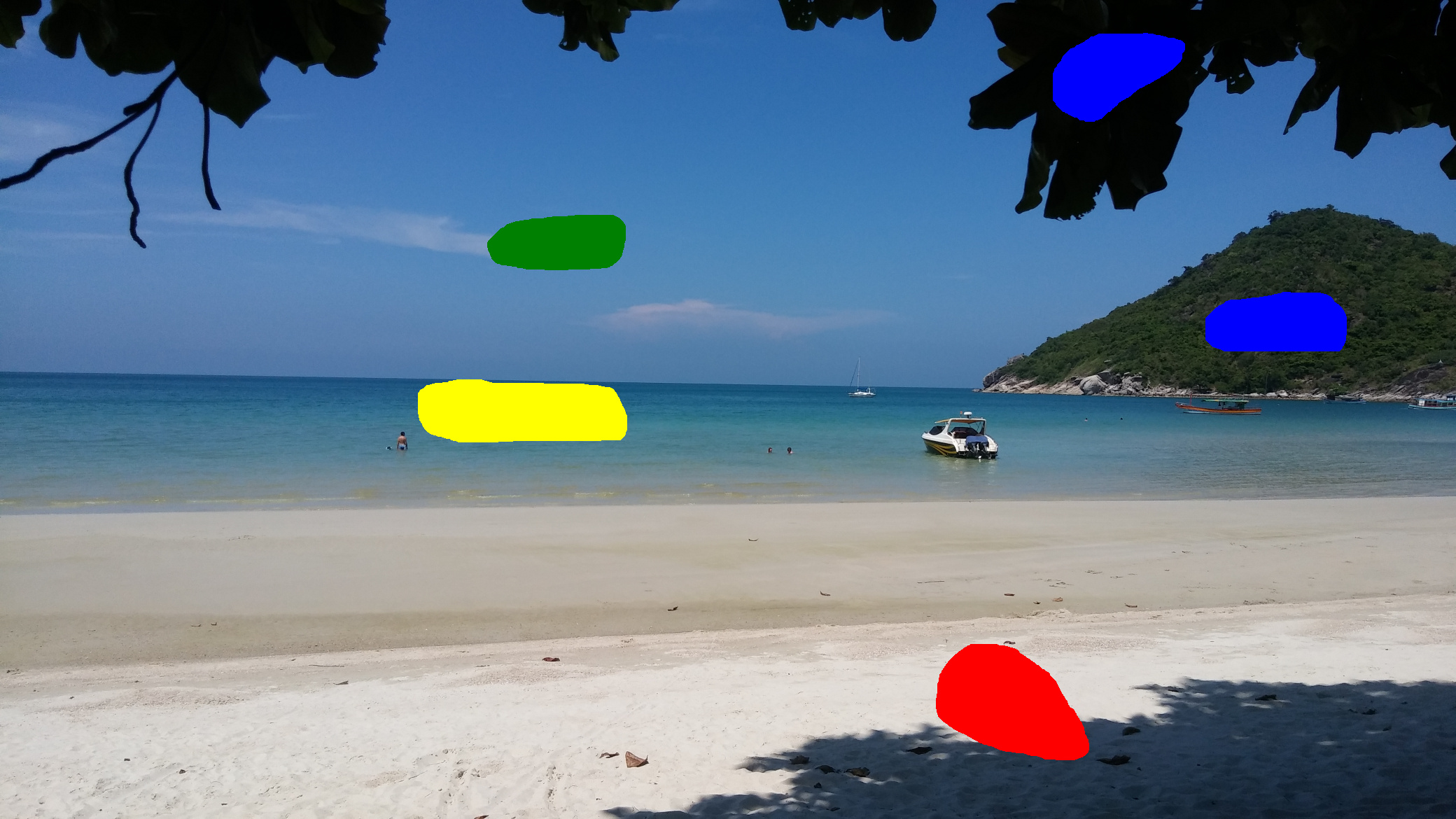}
\end{center}
\caption{Left: Test image $1$ ($4160 \times 2340$ pixels) for image segmentation. Right: Pre-labeled pixels.}
 \label{fig:beach_image}
\end{figure}

Diffuse interface methods are heavily used in materials science and beyond, see e.g.\ \cite{onuki2002phase,wheeler1992phase,wodo2012modeling,bergermann2019modeling} and the references therein. They offer an efficient and flexible way to model phase separation with one of the most prominent models being the Allen--Cahn equation~\cite{AllenCahn} that originally describes the evolution of a binary liquid over time. The Allen--Cahn equation is derived as the $L^2$ gradient flow of the Ginzburg--Landau energy functional

\begin{equation}\label{GinzburgLandau}
 E(u) = \int \frac{\epsilon}{2} | \nabla u |^{2} \mathrm{d}\boldsymbol{x} + \int \frac{1}{\epsilon} \psi (u) \mathrm{d}\boldsymbol{x}, 
\end{equation}
where we obtain
\begin{equation}\label{AllenCahn}
 \frac{\partial u}{\partial t} = - \nabla E(u) = \epsilon \Delta u - \frac{1}{\epsilon} \psi ' (u).
\end{equation}
Here, $u$ and $-u$ are defined on a physical domain~$\Omega$ and describe the two (liquid) components. Furthermore, $t$ is the time variable and $\epsilon>0$ is a (typically small) parameter which influences the width of the interface regions where the transition from one pure phase to the other happens. The gradient term represents the Dirichlet energy that penalizes the length of the interface and $\psi$ is a suitable potential function with minima at the pure phases.
Standard solution methods for partial differential equations, such as the finite element method or the finite difference method, can be used to solve \eqref{AllenCahn}, given suitable initial conditions as well as boundary conditions.\footnote{For details on \eqref{AllenCahn} as well as binary semi-supervised Allen--Cahn classification on graphs we refer to \Cref{sec:binary_allen_cahn}.}

The Allen--Cahn equation has been adapted to binary semi-supervised classification on graphs, see \cite{BertozziFlenner2012DiffuseHighdimClassif,calatroni2017graph}. Most importantly, the spatial domain~$\Omega$ is replaced by the graph domain, and in particular, this means that $\Omega$ is identified with the set of vertices $x_{i} \in \mathcal{V}$ of a graph~$\mathcal{G}$.
Moreover, a data fidelity term is added to the Ginzburg--Landau energy functional~\eqref{GinzburgLandau} to include the correct classification of the a priori labeled data as an additional objective.

Furthermore, the method has been extended to the multiclass case, see e.g.\ \cite{Garcia2014MulticlassDataSegm,BoschKlamtStoll2018DiffuseInterface,mercado2019node} as well as \Cref{fig:beach_evolution,fig:beach_image} for an image segmentation example.
For the general case of $m\geq 2$ classes, 
the vector $u \in \mathbb{R}^n$ of the phase-field description on the finite set of vertices $\mathcal{V}=\{x_{i}\}_{i=1}^n$ is identified with a matrix $\boldsymbol{U}\in\mathbb{R}^{n\times m}$ containing one row of $m$ entries per vertex~$x_i$.
The goal is to identify vertices $x_i$ with the $j$-th class whenever the $j$-th entry in the corresponding row $\boldsymbol{u}_i^\top$ of $\boldsymbol{U}$ is largest for the class ID $j\in\{1,\ldots,m\}$.
In order to incorporate the known class information of a priori labeled data,
the data fidelity term
\begin{equation*}
\frac{1}{2} \sum_{i=1}^{n} \omega(x_{i}) \| \boldsymbol{f}_{i}-\boldsymbol{u}_{i} \|_{\ell^2}^{2}
\end{equation*}
is added to the Ginzburg--Landau energy functional~\eqref{GinzburgLandau} in the multiclass case, where $\omega(x_i)$ is a penalty parameter that is equal to the constant $\omega_{0}\gg 0$ for labeled vertices~$x_i$ and $0$ for unlabeled vertices.
For the class information of the known labels, one-hot encoding is used, i.e., the vectors $\boldsymbol{f}_{i}$ are set to the $j$-th unit vector $\boldsymbol{e}_j\in\mathbb{R}^m$ for labeled data, 
\begin{equation}\label{def:f_i}
 \boldsymbol{f}_{i} := 
 \begin{cases}
  \boldsymbol{e}_j & \text{if node $x_i$ is of class $j$,} \\
  \boldsymbol{0} & \text{if class of node $x_i$ is unknown.}
 \end{cases}
\end{equation}

The choice of $\omega_0$ controls the trade-off between the classical Ginzburg--Landau energy and the least squares data fidelity term. Choosing $\omega_0$ too small bears the risk of underfitting while choosing it too large may cause overfitting to the pre-classified data.

In addition, the Dirichlet energy term $\int \frac{\epsilon}{2} | \nabla u |^{2} \mathrm{d}\boldsymbol{x}$ in the Ginzburg--Landau energy functional~\eqref{GinzburgLandau} is replaced by
$\frac{\epsilon}{2} \text{trace}(\boldsymbol{U}^\top \boldsymbol{L}_{\mathrm{sym}} \boldsymbol{U})$, which is motivated by \cite[Section~5.2]{luxburg}.
Moreover, the multi-well potential
\begin{equation}
\psi (\boldsymbol{u}_{i}) := \prod_{j=1}^{m} \frac{1}{4} \| \boldsymbol{u}_{i} - \boldsymbol{e}_{j} \|^{2}_{\ell^1} = \prod_{j=1}^{m} \frac{1}{4} \left( \sum_{l=1}^{n} |\boldsymbol{u}_{il} - \delta_{jl}| \right)^{2}
\label{equ:def_multiwell}
\end{equation}
is used with minima of $0$ in the corners of the Gibbs simplex 
\begin{equation}\label{gibbssimplex}
\Sigma^{m} := \left\{  (s_{1}, \dots , s_{m}) \in [0,1]^{m} \colon \sum_{j=1}^{m} s_{j} = 1 \right\},
\end{equation}
cf.~\cite{Garcia2014MulticlassDataSegm}, and the integral is replaced by a finite sum over the vertices $x_i$, $i=1,\ldots,n$.
The locations of these minima correspond to the one-hot encoding~\eqref{def:f_i} of the known class labels~$\boldsymbol{f}_{i}$ and model the goal that each row $\boldsymbol{u}_i^\top$ of the solution~$\boldsymbol{U}$ should be close to one of the unit vectors~$\boldsymbol{e}_j^\top$, $j\in\{1,\ldots,m\}$. Note that for the binary case $m=2$, the multi-well potential~\eqref{equ:def_multiwell} corresponds to the double-well potential~\eqref{defpsi}.

With the modifications discussed above, the discretized Ginzburg--Landau functional for the multiclass case becomes
\begin{equation}\label{equ:tilde_E_U}
\tilde{E}(\boldsymbol{U}) = \frac{\epsilon}{2} \text{trace}(\boldsymbol{U}^\top \boldsymbol{L}_{\mathrm{sym}} \boldsymbol{U}) + \frac{1}{2 \epsilon} \sum_{i=1}^{n} \left(\prod_{l=1}^{m} \frac{1}{4} \| \boldsymbol{u}_{i} - \boldsymbol{e}_{l} \|^{2}_{\ell^1} \right) + \frac{1}{2} \sum_{i=1}^{n} \omega(x_{i}) \| \boldsymbol{f}_{i}-\boldsymbol{u}_{i} \|_{\ell^2}^{2}
\end{equation}
and forms the basis for the semi-supervised classification technique considered in this work, as it can be viewed as its loss function. 
The potential term in this case slighty varies in contrast to the binary case as the scaling changes from $\frac{1}{\epsilon}$ to $\frac{1}{2\epsilon}$ and we stick to this formulation as this is used throughout the literature for the multiclass case. 

\begin{sloppypar}
For the solution of the Allen--Cahn equation~\eqref{AllenCahn}, a numerical scheme called convexity splitting \cite{eyre,Eyre97anunconditionally} is commonly applied, see e.g.\ \cite{mercado2019node,BoschKlamtStoll2018DiffuseInterface,BertozziFlenner2012DiffuseHighdimClassif,Garcia2014MulticlassDataSegm}, where $E(u)$ is split up into a convex part $E_1$ and a concave part $-E_2$ so that
$
 E(u) = E_1(u)-E_2(u).
$ 
The convex part $E_1$ is then treated implicitly to allow for numerical stability while the concave part $-E_2$ is treated explicitly.
For the general multiclass case $m\geq 2$, one possible splitting
$\tilde{E}(\boldsymbol{U})=\tilde{E}_1(\boldsymbol{U})-\tilde{E}_2(\boldsymbol{U})$ reads
\begin{align}
\label{equ:tilde_E_U_1}
 \tilde{E}_1(\boldsymbol{U}) &= \frac{\epsilon}{2} \text{trace}(\boldsymbol{U}^\top \boldsymbol{L}_{\mathrm{sym}} \boldsymbol{U}) + \frac{c}{2} \text{trace}(\boldsymbol{U}^\top \boldsymbol{U}), \\
 \tilde{E}_2(\boldsymbol{U}) &= \frac{c}{2} \text{trace}(\boldsymbol{U}^\top \boldsymbol{U}) - \frac{1}{2 \epsilon} \sum_{i=1}^{n} \left(\frac{1}{4} \prod_{l=1}^{m} \| \boldsymbol{u}_{i} - \boldsymbol{e}_{l} \|^{2}_{\ell^1} \right) - \frac{1}{2} \sum_{i=1}^{n} \omega(x_{i}) \| \boldsymbol{f}_{i}-\boldsymbol{u}_{i} \|_{\ell^2}^{2},\\
\nonumber
 &= \sum_{i=1}^{n}\frac{c}{2} (\boldsymbol{u}_i^\top \boldsymbol{u}_i) - \frac{1}{2 \epsilon}  \left(\frac{1}{4} \prod_{l=1}^{m} \| \boldsymbol{u}_{i} - \boldsymbol{e}_{l} \|^{2}_{\ell^1} \right) - \frac{1}{2} \omega(x_{i}) \| \boldsymbol{f}_{i}-\boldsymbol{u}_{i} \|_{\ell^2}^{2},
\end{align}
where a productive zero is inserted into $\tilde{E}(\boldsymbol{U})$ by adding and subtracting the convex function $\frac{c}{2} \text{trace}(\boldsymbol{U}^\top \boldsymbol{U})$ with a constant $c>0$ that ensures the strict convexity of $\tilde{E}_2$, cf.~\cite{Garcia2014MulticlassDataSegm}.
The resulting scheme is then given by 
\begin{equation}\label{convexitysplittingschememulti}
 \frac{ \boldsymbol{U}^{l+1} - \boldsymbol{U}^{l}}{\Delta t} = - \epsilon \boldsymbol{L}_{\mathrm{sym}}\boldsymbol{U}^{l+1} - c \boldsymbol{U}^{l+1} + c \boldsymbol{U}^{l} - \frac{1}{2 \epsilon} \boldsymbol{\mathcal{T}}^{l} - \boldsymbol{\omega} \, (\boldsymbol{U}^{l}-\boldsymbol{F}).
\end{equation}
Here the derivative of the term $\frac{1}{2 \epsilon} \sum_{i=1}^{n} (\prod_{l=1}^{m} \frac{1}{4} \| \boldsymbol{u}_{i} - \boldsymbol{e}_{l} \|^{2}_{\ell^1} )$ is given by the matrix $\boldsymbol{\mathcal{T}}(\boldsymbol{U}) \in \mathbb{R}^{n \times m}$ with the entries 
\begin{equation}\label{def:T_ij}
 \mathcal{T}_{ij}(\boldsymbol{U}):=
 \sum_{q=1}^{m} \frac{1}{2} (1- 2 \delta_{jq}) \| \boldsymbol{u}_{i} - \boldsymbol{e}_{q} \|_{\ell^1} \prod_{\substack{p = 1\\p \neq q}}^{m} \frac{1}{4} \| \boldsymbol{u}_{i} - \boldsymbol{e}_{p} \|^{2}_{\ell^1},
\end{equation}
$\boldsymbol{\mathcal{T}}^{l}:=\boldsymbol{\mathcal{T}}(\boldsymbol{U}^l)$,
$\boldsymbol{\omega}:=\diag\left(\omega(x_{i})_{i=1}^n\right)$, and $\boldsymbol{F}:=(\boldsymbol{f}_i^\top)_{i=1}^n\in\mathbb{R}^{n\times m}$ contains the known label information, cf.~\eqref{def:f_i}. 
All of this leads to solving the equation
\begin{equation}
\label{equ:U_lp1}
 \boldsymbol{U}^{l+1} = \big[ \underbrace{(1+c(\Delta t)) \boldsymbol{I} + \epsilon (\Delta t) \boldsymbol{L}_{\mathrm{sym}}}_{:=\boldsymbol{B}} \big]^{-1} \left( (1+c(\Delta t))\boldsymbol{U}^{l} - \frac{\Delta t}{2 \epsilon} \boldsymbol{\mathcal{T}}^{l} - (\Delta t) \, \boldsymbol{\omega} \, (\boldsymbol{U}^{l} - \boldsymbol{F}) \right),
\end{equation}
which is made computationally efficient using a projection onto the dominating eigenspace of $\boldsymbol{L}_{\mathrm{sym}}$, see
\Cref{sec:convexitysplittingmulticlass} for details.
\end{sloppypar}

For the initialization of $\boldsymbol{U}^{0}$, we set $\boldsymbol{u}_{i}:=\boldsymbol{e}_{j} \in \mathbb{R}^{m}$ for the pre-labeled vertices $x_{i}$ where $j$ is the corresponding class. For the vertices with unknown labels, \cite[Algorithm 1]{Garcia2014MulticlassDataSegm} suggests to use randomized initial conditions $\boldsymbol{U}^{0}$, which are then scaled to the Gibbs simplex. Especially in unsupervised classification problems, this initialization strategy can be beneficial, cf.\ \cite[Section~5.5]{tudisco2018community} for a detailed discussion. However, interpreting the vector entries of $\boldsymbol{u}_i \in \Sigma^m$ as empirical probabilities for class affiliation, a random initialization bears the risk of initially assigning high probabilities to wrong classes, which potentially reduces the classification accuracy of the whole method. Having no a priori information about the unlabeled nodes, we therefore propose to initialize the respective rows of $\boldsymbol{U}^0$ with uniform empirical probabilities, i.e.\ $\boldsymbol{u}_{i}:=\frac{1}{m} \boldsymbol{1} \in \mathbb{R}^{m}$. Numerical experiments with our semi-supervised classifier (not included in \Cref{sec:numerics}) underpin this heuristic as classification results consistently degrade with an increase of randomization in the initialization.

Note, that the presented method requires the choice of several hyper-parameters. While the Allen--Cahn parameters in the ranges $\epsilon \in [5 \cdot 10^{-3}, 0.5],\ \omega_0 \in [10^3, 10^4],\ c \in [\omega_0+1/\epsilon, \omega_0+3/\epsilon]$ and $\Delta t \in [10^{-2}, 1]$ were experimentally found not to have a major influence on classification results, the number $k$ of eigenpairs of the Laplacian operator and the scaling parameter $\sigma$ in the Gaussian kernel have to be carefully chosen. We employed a parameter-grid search for both parameters for each data set presented in the numerical experiments in \Cref{sec:numerics}.

A related technique involving fewer hyper-parameters is based on the Merriman--Bence--Osher (MBO) scheme where the nonlinear term of the PDE is dropped and a tresholding procedure is added to the method, cf.\ e.g.~\cite{merkurjev2013mbo,Garcia2014MulticlassDataSegm,bertozzi2016diffuse} and references therein. This approach could potentially reduce the computation time per iteration as the nonlinearity does not need to be evaluated. A naive parameter selection did not yield improved performance for this scheme compared to our approach when applied on the WebKB data set from \Cref{sec:numerics:multilayer}. We leave a detailed investigation to future research. Moreover, we point out that all techniques suggested in \Cref{sec:eigenpairs_PKSM,sec:nfft_fastsum,sec:feature_grouping} should also be applicable for the MBO scheme.

\section{Graph Allen--Cahn on multilayer graphs for multiclass problems}\label{sec:graph_allen_cahn_multilayer} 

In this section, we extend the graph-based multiclass Allen--Cahn approach from \Cref{sec:graph_allen_cahn_multiclass} to multilayer graphs. For this, we employ the concept of the power mean Laplacian in order to combine the information of all graph layers, see \Cref{sec:graphs_and_multilayer}.
Formally, we replace the graph Laplacian matrix $\boldsymbol{L}_{\mathrm{sym}}$ in~\eqref{equ:tilde_E_U}
by the power mean Laplacian $\boldsymbol{L}_p$ from~\eqref{equ:power_mean_laplacian} for $p>0$ (corresponding to a shift $\delta=0$)
and by the shifted version $\boldsymbol{L}_{p,\delta}$ from~\eqref{equ:power_mean_laplacian_shifted} with $\delta=\log (1+|p|)$ for $p<0$.
This causes changes in the term $\tilde{E}_1(\boldsymbol{U})$ in~\eqref{equ:tilde_E_U_1}, and correspondingly, $\boldsymbol{L}_{\mathrm{sym}}$ is replaced in the iteration formula~\eqref{equ:U_lp1} yielding
\begin{equation*}
 \boldsymbol{B}=
 \begin{cases}
   (1+c(\Delta t)) \boldsymbol{I} + \epsilon (\Delta t) \boldsymbol{L}_p & \text{for } p>0, \\
   (1+c(\Delta t)) \boldsymbol{I} + \epsilon (\Delta t) \boldsymbol{L}_{p,\delta} & \text{for } p<0.  
 \end{cases}
\end{equation*}
Analogously to the single layer case, we utilize an eigendecomposition $\boldsymbol{\Phi} \boldsymbol{\Lambda} \boldsymbol{\Phi}^\top$ of $\boldsymbol{L}_p$ and $\boldsymbol{L}_{p,\delta}$ for $p>0$ and $p<0$, respectively,
where $\boldsymbol{\Lambda}\in\mathbb{R}^{n\times n}$ and $\boldsymbol{\Phi}\in\mathbb{R}^{n\times n}$.
The eigendecomposition will then be approximated by a truncated version $\boldsymbol{\Lambda}_k\in\mathbb{R}^{k\times k}$, $\boldsymbol{\Phi}_k\in\mathbb{R}^{n\times k}$ using the $k$ smallest eigenvalues. In order to compute this eigeninformation, we use the Lanczos method, which relies on matrix-vector products with~$\boldsymbol{L}_p$ and~$\boldsymbol{L}_{p,\delta}^p$.
For the case $p=1$, which is the computationally most efficient one as it does not involve matrix powers, we compute the eigenpairs belonging to the $k$ largest eigenvalues of~\eqref{equ:I-L_1} as described in detail in \Cref{sec:eigenpairs_PKSM}.
Otherwise, for the case $p<0$, we use the relation 
\begin{equation*}
 \boldsymbol{L}_{p,\delta}^p = \frac{1}{T} \sum_{t=1}^T (\boldsymbol{L}_{\mathrm{sym},\delta}^{(t)})^p
\end{equation*}
to compute the eigenpairs belonging to the $k$ largest eigenvalues $\lambda_i^p, i \in \{ 1, \dots , k\}$ of the $p$th power of the shifted power mean Laplacian $\boldsymbol{L}_{p,\delta}^p$ again using the Lanczos method. Within the Lanczos process, we employ the Polynomial Krylov Subspace Method \cite{pmlr-v84-mercado18a} in order to approximate the matrix powers of the single layer graph Laplacians $(\boldsymbol{L}_{\mathrm{sym},\delta}^{(t)})^p$ as described in \Cref{sec:eigenpairs_PKSM}. 

The graph Allen--Cahn equation can then be solved using the iteration\footnote{This can be obtained from \eqref{equ:U_lp1} as shown in the detailed derivation in \Cref{sec:convexitysplittingmulticlass}.}
\begin{equation*}
\boldsymbol{V}^{l+1} = \big[ (1+c(\Delta t)) \boldsymbol{I} + \epsilon (\Delta t) \boldsymbol{\Lambda}_k \big]^{-1} \boldsymbol{\Phi}_k^\top \left( (1+c(\Delta t))\boldsymbol{U}^{l} - \frac{\Delta t}{2 \epsilon} \boldsymbol{\mathcal{T}}^{l} - (\Delta t) \, \boldsymbol{\omega} \, (\boldsymbol{U}^{l} - \boldsymbol{F}) \right),
\end{equation*}
since it only depends on the eigendecomposition of the graph Laplacian, the known label information~$\boldsymbol{F}$, and the previous iterate $\boldsymbol{U}^{l}$ ($\boldsymbol{\mathcal{T}}^l$ is a function of $\boldsymbol{U}^{l}$). As before, we project each row of the result $\boldsymbol{\tilde{U}}^{l+1}:=\boldsymbol{\Phi}_k\boldsymbol{V}^{l+1}\in\mathbb{R}^{n\times m}$ of each iteration~$l$ back to the Gibbs simplex $\Sigma^{m}$ by the method~\cite{chen2011projection} and use these projected values as the input $\boldsymbol{U}^{l+1}$ for the next iteration $l+1$.

Finally, we obtain the method summarized in \Cref{alg::multilayer_graph_allen_cahn_multiclass}. The algorithm takes the multilayer graph Laplacian and the known label information as inputs. \Cref{alg::multilayer_graph_allen_cahn_multiclass} outputs the scores for the class affiliations of each node $x_i\in\mathcal{V}$ of the multilayer graph.
Based on these scores, we predict the class by a majority vote, i.e., we take the row-wise maximum of the output matrix $\boldsymbol{U}^l$, where ties are broken by prioritizing the class with the lowest class ID.

Note that  \Cref{alg::multilayer_graph_allen_cahn_multiclass} does not employ a fast technique for the case $1 \neq p > 0$. The reason is that this would require the smallest, i.e.\ informative eigenvalues, of the power mean Laplacian,  which due to the nature of typical iterative eigenvalue solvers need to be computed using the inverse of the matrix, cf.\ e.g.~\cite[Sec.~7.6.1]{golub2012matrix}. This would then add another Lanczos iteration on top of the existing ones, which would further increase the algorithmic complexity. It would be possible to investigate the use of the Rayleigh-Chebyshev iteration \cite{anderson2010rayleigh} in combination with the power mean Laplacian but the results presented in \cite{pmlr-v84-mercado18a} as well as our experiments in \Cref{sec:numerics:SBM} indicate that positive $p$'s tend to achieve worse classification results compared to the case $p<0$ for which we present efficient numerical methods.

\begin{algorithm}[htb!]
\caption{Computation of the multiclass scores for the class affiliations of each node of the multilayer graph using a graph Allen--Cahn type method.}
\label{alg::multilayer_graph_allen_cahn_multiclass}
\vspace{0.5em}
  \begin{tabular}{p{1.4cm}p{4.5cm}p{8.35cm}}
    Input:
                & $\boldsymbol{L}_{\mathrm{sym}}^{(t)}\in\mathbb{R}^{n\times n}$, \newline $t=1,\ldots,T$ & graph Laplacian matrix for each layer or function realizing matrix-vector multiplication of graph Laplacian matrix with a vector, \\[0.4em]
                & $\boldsymbol{F}:=(\boldsymbol{f}_i^\top)_{i=1}^n\in\{0,1\}^{n\times m}$  & known label information as per~\eqref{def:f_i}. \\
\vspace{1em}
  \end{tabular}
  \begin{tabular}{p{2cm}p{10cm}}
   Parameters: & $p$, $k$, $\epsilon$, $\omega_0$, $c$, $\Delta t$, \verb|tolerance|, \verb|max_iter| \\
  \end{tabular}
\vspace{1em}
  \begin{enumerate}
    \item If $p=1$, then compute eigenpairs $\boldsymbol{\tilde{\Lambda}}_k,\boldsymbol{\Phi}_k$ of $\boldsymbol{I} - \boldsymbol{L}_1$ belonging to the $k$ largest eigenvalues, using the Lanczos method. Set $\boldsymbol{\Lambda}_k := \boldsymbol{I} - \boldsymbol{\tilde{\Lambda}}_k$. \newline
    Otherwise, for $p<0$, using the Lanczos method in combination with the PKSM~\cite{pmlr-v84-mercado18a}, compute the $k$-largest eigenvalues $\lambda_1^p, \dots , \lambda_k^p$ of~$\boldsymbol{L}_{p,\delta}^p$ with its eigenvectors, using $\delta:=\log (1+|p|)$. Collect the eigenvectors in~$\boldsymbol{\Phi}_k$. Obtain $\boldsymbol{\Lambda}_k := \diag (\lambda_1,\ldots,\lambda_k)$.
    \item Compute $\boldsymbol{Z}:=\big[ (1+c(\Delta t)) \boldsymbol{I} + \epsilon (\Delta t) \boldsymbol{\Lambda}_k \big]^{-1} \boldsymbol{\Phi}_k^\top$.
    \item Initialize $\boldsymbol{U}_0:=\big((\boldsymbol{u}_i^0)^\top\big)_{i=1}^n \in \mathbb{R}^{n\times m}$ with $\boldsymbol{u}_{i}^0:=\boldsymbol{e}_{j} \in \mathbb{R}^{m}$ for the pre-labeled vertices~$x_{i}$ where $j$ is the corresponding class. Otherwise, use $\boldsymbol{u}_{i}^0:=\frac{1}{m} \boldsymbol{1} \in \mathbb{R}^{m}$ for unlabeled~$x_{i}$. \newline Initialize $l:=0$.
    \item \textbf{do}
    \begin{enumerate}
      \item Compute the matrix $\boldsymbol{\mathcal{T}}^l=\boldsymbol{\mathcal{T}}(\boldsymbol{U}^l)$ with entries~\eqref{def:T_ij}.
      \item Compute $\boldsymbol{V}^{l+1}:= \boldsymbol{Z} \left( (1+c(\Delta t))\boldsymbol{U}^{l} - \frac{\Delta t}{2 \epsilon} \boldsymbol{\mathcal{T}}^{l} - (\Delta t) \, \boldsymbol{\omega} \, (\boldsymbol{U}^{l} - \boldsymbol{F}) \right)$.
      \item Project rows of~$\boldsymbol{\tilde{U}}^{l+1} := \boldsymbol{\Phi}_k\boldsymbol{V}^{l+1}\in\mathbb{R}^{n\times m}$ to Gibbs simplex~$\Sigma^m$ using~\cite{chen2011projection} and obtain new iterate~$\boldsymbol{U}^{l+1}$.
      \item Calculate \verb|relative_change| $:=$ $\max_{i=1,\ldots,n} \|\boldsymbol{u}_i^{l+1} - \boldsymbol{u}_i^l\|^2 / \max_{i=1,\ldots,n} \|\boldsymbol{u}_i^{l+1}\|^2$.
      \item $l\leftarrow l+1$.
     \end{enumerate}
    \item[] \textbf{while} \verb|relative_change| $>$ \verb|tolerance| and $l$ $<$ \verb|max_iter|
  \end{enumerate}
\vspace{1em}
  \begin{tabular}{p{1.4cm}p{5cm}p{7.85cm}}
    Output: & $\boldsymbol{U}^{l}:=\big((\boldsymbol{u}_i^{l})^\top\big)_{i=1}^n\in [0,1]^{n\times m}$ & scores for class affiliation of each node $x_i$, $\boldsymbol{u}_i^{l}\in\Sigma^m$. \\
    \cmidrule{1-3}
 \end{tabular}
\end{algorithm}

When the weight matrix~$\boldsymbol{W}^{(t)}$, which encodes node similarities in each layer $t\in\{1,\ldots,T\}$, is associated with a suitable\footnote{for example, when $K$ is rotationally invariant and smooth} kernel function $K^{(t)}\colon\mathbb{R}^d\rightarrow\mathbb{R}_{\geq 0}$ as discussed in \Cref{sec:nfft_fastsum}, we can substantially accelerate the computation of the required eigenpairs of the power mean Laplacian by using the NFFT-based fast summation in the case $d\leq 3$\footnote{The case $d \geq 4$ is discussed in the next section.}. This is true even when each weight matrix~$\boldsymbol{W}^{(t)}$ is densely populated, as the computational complexity of one matrix-vector multiplication with~$\boldsymbol{W}^{(t)}$ is reduced from $\mathcal{O}(n^2)$ to $\mathcal{O}(n)$ with the help of \Cref{alg::nfft_fastsum}.

\section{Higher dimensional data and feature grouping}\label{sec:feature_grouping}

In this section, we consider the case $d \geq 4$ for which \Cref{alg::nfft_fastsum} from \Cref{sec:nfft_fastsum} would be computationally too expensive and thus, would no longer be directly applicable within a reasonable time frame. This section introduces a framework that makes the NFFT-based methods efficient again. For these higher dimensional data sets, the formation of the weight matrix $\boldsymbol{W}$ can be accomplished with a computational complexity of $\mathcal{O}(d\,n^2)$ when the evaluation of the kernel function is $\mathcal{O}(d)$. The memory requirement of this approach is at most $n^2$ and each matrix-vector product can be realized in a computational complexity of $\mathcal{O}(n^2)$.
When $\boldsymbol{W}$ cannot be stored explicitly, each row of $\boldsymbol{W}$ can also be assembled on-the-fly whenever a matrix-vector multiplication with $\boldsymbol{W}$ is performed, increasing the computational complexity from $\mathcal{O}(n^2)$ to $\mathcal{O}(d\,n^2)$ (still assuming $\mathcal{O}(d)$ for each evaluation of the kernel function).  For large $n$, however, the explicit storage or on-the-fly construction of the matrices quickly becomes infeasible for practical computations and an alternative approach is required.

Incorporating the NFFT-based fast summation from \Cref{sec:nfft_fastsum} adapts the computational complexity from $\mathcal{O}(n^2)$ to $\mathcal{O}\big((\texttt{m}_\text{NFFT})^{d}\,n + d \, 2^d N^{d} \log N\big)$. This linear dependence on $n$ enables the treatment of large data sets. In order to decrease the exponential dependence of this complexity on the spatial dimension~$d$, we propose a feature grouping approach for $d \geq 4$. The idea is to decompose the $d$-dimensional feature space into $T$ subspaces of (not necessarily equal) dimensions $d^{(t)} \leq 3$, so that
\begin{equation*}
 d = \sum_{t=1}^T d^{(t)}.
\end{equation*}

We interpret the data $\boldsymbol{X}^{(t)}$ in each subspace as one graph layer $\mathcal{G}^{(t)}$ in a multilayer graph as introduced in \Cref{sec:graphs_and_multilayer}. Defining a weight function~$w^{(t)}$ for each layer enables the formation of the weight matrices $\boldsymbol{W}^{(t)}$, the degree matrices $\boldsymbol{D}^{(t)}$ and finally the individual layer graph Laplacians $\boldsymbol{L}_{\mathrm{sym}}^{(t)}$. 

While there is no theory on an optimal feature grouping yet, numerical experiments show, that grouping ``suitable'' features together leads to much higher accuracies in classification tasks. 
Considering the example of an RBG image, the best image segmentation results can be obtained using a very reasonable grouping of the $5$-dimensional feature space consisting of the RGB-values and the $x$- and $y$-pixel coordinates into one layer containing the RGB color information and a second layer containing the spatial $xy$ information of the pixel\footnote{We ran several numerical tests with different groupings. The results are not included in this manuscript, but can be reproduced using our code linked in \Cref{sec:numerics}.}, cf.~\cite{tomasi1998bilateral}. 
This approach combined with the method presented in \Cref{sec:feature_grouping} will allow us to consider large image data sets in our numerical tests in \Cref{sec:numerics:image}, applying \Cref{alg::multilayer_graph_allen_cahn_multiclass} to a 10 megapixel image which then corresponds to a two layer graph with 10 million nodes per layer.

The feature grouping approach thus gives rise to a new class of multilayer graphs, where the information from the different graph layers $\mathcal{G}^{(t)}$ is merged together again using the individual layer graph Laplacians $\boldsymbol{L}_{\mathrm{sym}}^{(t)}$ in the power mean Laplacian defined in \eqref{equ:power_mean_laplacian_shifted}. The combination of this approach with the ideas from \Cref{sec:eigenpairs_PKSM} allows a reduction of the computational complexity of matrix-vector products with the power mean Laplacian $\boldsymbol{L}_{p,\delta}$ of large high-dimensional data sets to approximately $\mathcal{O}(d\,n)$. The power mean Laplacian is then used in the Allen--Cahn classification scheme\footnote{This is another $\mathcal{O}(n)$ operation which depends on the termination of the eigeninformation computations.} as described in \Cref{sec:graph_allen_cahn_multilayer}.

\section{Numerical experiments}\label{sec:numerics}

In this section, we present the numerical results for the various methods and approaches discussed in the previous sections.\footnote{In addition, \Cref{sec:numerics:SBM} presents numerical experiments on synthetic stochastic block model data.} The corresponding \textsc{MATLAB} code is available at \url{https://www.tu-chemnitz.de/mathematik/wire/codes.php}. All runtime measurements were performed on a laptop computer with 16 GB RAM and an Intel Core i5-8265U CPU with 4 $\times$ 1.60--3.90~GHz cores. We use the shift $\delta=\log(1+|p|)$ for $p<0$ and $\delta=0$ for $p>0$ for the power mean Laplacian defined in \eqref{equ:power_mean_laplacian_shifted} throughout this section.

\subsection{Small multilayer data sets}\label{sec:numerics:multilayer}

We start by classifying
some small real world example data sets with an inherent multilayer structure that still permit the explicit formation of the power mean Laplacian.
Here we consider the data sets from \cite[Section~6]{mercado2019generalized},
where the three data sets Citeseer~\cite{lu2003link}, Cora~\cite{mccallum2000automating} and WebKB~\cite{craven1998learning} have one layer,
the three data sets 3sources \cite{liu2013multi}, BBCS~\cite{greene2009matrix} and Wikipedia~\cite{wiki} have three layers,
the data set BBC~\cite{greene2005producing} has four layers, and the data set UCI~\cite{UCIBreukelen1998HandwrittenDR} six layers. For each multilayer data set, the feature space dimensionality varies across the graph layers ranging between $d^{(t)}=6$ and $d^{(t)}=4\,684$. As $n$ is sufficiently small in all examples, we are in situation (i) of the enumeration in \Cref{sec:graphs_and_multilayer} and explicitly assemble and store all matrices and do not apply feature grouping and the NFFT.

We chose these data sets as they are readily available, and we can compare our new results with the multilayer semi-supervised learning (SSL) approach presented in \cite[Section~3]{mercado2019generalized}.
This approach uses the power mean Laplacian for a generalized Tikhonov type regularization, also called ridge regression, solving the optimization problem
\begin{equation}
(u_{i,j})_{i=1}^n = \text{arg} \min_{\boldsymbol{y}_j \in \mathbb{R}^n} \| \boldsymbol{y}_j - (f_{i,j})_{i=1}^n \|^2 + \lambda \boldsymbol{y}_j^\top\boldsymbol{L}\boldsymbol{y}_j
\label{equ:SSL_Pedro}
\end{equation}
for each class $j\in\{1,\ldots,m\}$, 
where $\boldsymbol{L}$ is either $\boldsymbol{L}_1$ or $\boldsymbol{L}_{p,\delta}$ with $p<0$,
and the label of a node $v_i$ is determined by $\text{arg} \max \{ u_{i,j} \}_{j=1}^m$.
In particular, the test code for \cite[Section~6]{mercado2019generalized} including the known label information for the data sets can be obtained from~\cite{mercado2019generalizedSoftware}, and we were able to exactly reproduce the corresponding results in \cite[Table~2]{mercado2019generalized} which presents the mean misclassification rates of 10 test runs with different random known label information. We denote these results by the prefix ``SSL'' in \Cref{tab:results_pedro}. In contrast to the MBO scheme we here solve a linear system and then only once perform a thresholding to obtain the classes.

\begin{table}
\begin{minipage}{0.49\textwidth}
\begin{tiny}
\begin{center}
3sources~\cite{liu2013multi} 
($n=169$, $T=3$, $m=6$; AC: $\sigma=5$, $k=16$)
\begin{tabular}{|l c c c c c c|}
\hline \hline
   known labels\rule[0em]{0em}{1em} & $1\%$ & $5\%$ & $10\%$ & $15\%$ & $20\%$ & $25\%$\\ \hline
   SSL $\boldsymbol{L}_1$\rule[0em]{0em}{1em} & $33.5$ & $23.9$ & $23.4$ & $20.1$ & $15.6$ & $14.6$ \\
   SSL $\boldsymbol{L}_{-1,\delta}$ & $\bm{28.4}$ & $\bm{20.0}$ & $21.8$ & $22.0$ & $17.2$ & $17.9$ \\
   SSL $\boldsymbol{L}_{-10,\delta}$\rule[-0.4em]{0em}{1em} & $40.9$ & $29.1$ & $21.9$ & $19.3$ & $14.8$ & $14.7$ \\
\hline
   AC $\boldsymbol{L}_1$\rule[0em]{0em}{1em} & $39.9$ &  $25.1$ & $\bm{17.3}$ & $15.6$ & $\bm{10.6}$ & $\bm{10.2}$ \\
   AC $\boldsymbol{L}_{-1,\delta}$ & $40.2$ & $25.3$ & $\bm{17.3}$ & $\bm{15.3}$ & $\bm{10.6}$ & $10.4$ \\
   AC $\boldsymbol{L}_{-10,\delta}$\rule[-0.4em]{0em}{1em} & $38.5$ & $25.1$ & $17.6$ & $\bm{15.4}$ & $\bm{10.7}$ & $10.6$  \\
   \hline
  \end{tabular}\\[1em]

  BBCS~\cite{greene2009matrix} 
($n=544$, $T=2$, $m=5$; AC: $\sigma=2$, $k=62$)
  \begin{tabular}{|l c c c c c c|}
  \hline \hline
   known labels\rule[0em]{0em}{1em} & $1\%$ & $5\%$ & $10\%$ & $15\%$ & $20\%$ & $25\%$\\ \hline
   SSL $\boldsymbol{L}_1$\rule[0em]{0em}{1em} & $29.9$ & $15.0$ & $13.5$ & $10.6$ & ~$8.7$ & ~$7.2$ \\
   SSL $\boldsymbol{L}_{-1,\delta}$ & $\bm{23.8}$ & $\bm{11.6}$ & ~$8.7$ & ~$\bm{6.3}$ & ~$\bm{5.8}$ & ~$\bm{5.1}$ \\
   SSL $\boldsymbol{L}_{-10,\delta}$\rule[-0.4em]{0em}{1em} & $48.7$ & $22.5$ & $14.2$ & ~$9.1$ & ~$7.8$ & ~$6.1$ \\
\hline
   AC $\boldsymbol{L}_1$\rule[0em]{0em}{1em} & $52.4$ & $14.5$ & ~$\bm{8.5}$ & ~$6.5$ & ~$\bm{5.8}$ & ~$\bm{5.0}$ \\
   AC $\boldsymbol{L}_{-1,\delta}$ & $52.3$ & $14.7$ & ~$\bm{8.6}$ & ~$6.5$ & ~$\bm{5.8}$ & ~$\bm{5.0}$ \\
   AC $\boldsymbol{L}_{-10,\delta}$\rule[-0.4em]{0em}{1em} & $52.1$ & $14.2$ & ~$\bm{8.6}$ & ~$\bm{6.4}$ & ~$6.0$ & ~$\bm{5.0}$  \\
   \hline
  \end{tabular}\\[1em]

  UCI~\cite{UCIBreukelen1998HandwrittenDR} 
($n=2\,000$, $T=6$, $m=10$; AC: $\sigma=10$, $k=98$) 
  \begin{tabular}{|l c c c c c c|}
  \hline \hline
   known labels\rule[0em]{0em}{1em} & $1\%$ & $5\%$ & $10\%$ & $15\%$ & $20\%$ & $25\%$\\ \hline
   SSL $\boldsymbol{L}_1$\rule[0em]{0em}{1em} & $31.3$ & $23.8$ & $18.7$ & $15.6$ & $14.4$ & $13.2$ \\
   SSL $\boldsymbol{L}_{-1,\delta}$ & $\bm{30.5}$ & $17.1$ & $13.8$ & $12.6$ & $12.3$ & $11.9$ \\
   SSL $\boldsymbol{L}_{-10,\delta}$\rule[-0.4em]{0em}{1em} & $57.0$ & $33.8$ & $23.7$ & $17.6$ & $15.3$ & $13.4$ \\
\hline
   AC $\boldsymbol{L}_1$\rule[0em]{0em}{1em} & $39.6$ & $\bm{11.6}$ & ~$\bm{8.7}$ & ~$\bm{7.9}$ & ~$7.1$ & ~$\bm{6.5}$ \\
   AC $\boldsymbol{L}_{-1,\delta}$ & $44.5$ & $\bm{11.5}$ & ~$\bm{8.6}$ & ~$\bm{8.0}$ & ~$\bm{6.9}$ & ~$\bm{6.6}$ \\
   AC $\boldsymbol{L}_{-10,\delta}$\rule[-0.4em]{0em}{1em} & $47.4$ & $12.3$ & ~$9.1$ & ~$8.1$ & ~$7.2$ & ~$6.9$  \\
   \hline
  \end{tabular}\\[1em]

  Cora~\cite{mccallum2000automating} 
($n=2\,708$, $T=1$, $m=7$; AC: $\sigma=2$, $k=85$)
   \begin{tabular}{|l c c c c c c|}
   \hline \hline
   known labels\rule[0em]{0em}{1em} & $1\%$ & $5\%$ & $10\%$ & $15\%$ & $20\%$ & $25\%$\\ \hline
   SSL $\boldsymbol{L}_1$\rule[0em]{0em}{1em} & $50.7$ & $38.2$ & $33.4$ & $31.2$ & $28.2$ & $25.6$ \\
   SSL $\boldsymbol{L}_{-1,\delta}$ & $\bm{43.2}$ & $\bm{31.8}$ & $\bm{24.5}$ & $\bm{21.1}$ & $\bm{18.8}$ & $\bm{17.2}$ \\
   SSL $\boldsymbol{L}_{-10,\delta}$\rule[-0.4em]{0em}{1em} & $62.0$ & $46.3$ & $35.4$ & $29.4$ & $25.2$ & $22.3$ \\
\hline
   AC $\boldsymbol{L}_1$\rule[0em]{0em}{1em} & $62.3$ & $43.1$ & $34.6$ & $32.1$ & $30.3$ & $29.3$ \\
   AC $\boldsymbol{L}_{-1,\delta}$ & $62.3$ & $43.1$ & $34.6$ & $32.1$ & $30.3$ & $29.3$ \\
   AC $\boldsymbol{L}_{-10,\delta}$\rule[-0.4em]{0em}{1em} & $62.3$ & $43.1$ & $34.6$ & $32.1$ & $30.3$ & $29.3$ \\
   \hline
   \end{tabular}

   \end{center}
   \end{tiny}
  \end{minipage}
  \begin{minipage}{0.49\textwidth}
  \begin{tiny}
  \begin{center}
  BBC~\cite{greene2005producing} 
($n=685$, $T=4$, $m=5$; AC: $\sigma=6$, $k=31$)
  \begin{tabular}{|l c c c c c c|}
  \hline \hline
   known labels\rule[0em]{0em}{1em} & $1\%$ & $5\%$ & $10\%$ & $15\%$ & $20\%$ & $25\%$\\ \hline
   SSL $\boldsymbol{L}_1$\rule[0em]{0em}{1em} & $31.3$ & $22.8$ & $17.4$ & $13.5$ & $10.2$ & $8.9$ \\
   SSL $\boldsymbol{L}_{-1,\delta}$ & $\bm{31.0}$ & $17.0$ & $11.5$ & $10.5$ & ~$9.2$ & $8.7$ \\
   SSL $\boldsymbol{L}_{-10,\delta}$\rule[-0.4em]{0em}{1em} & $51.6$ & $26.9$ & $16.6$ & $12.8$ & $10.3$ & $9.5$ \\
\hline
   AC $\boldsymbol{L}_1$\rule[0em]{0em}{1em} & $41.9$ &  $\bm{12.9}$ & ~$8.9$ & ~$\bm{7.6}$ & ~$\bm{7.0}$ & $\bm{6.1}$ \\
   AC $\boldsymbol{L}_{-1,\delta}$ & $41.9$ & $\bm{13.0}$ & ~$\bm{8.8}$ &  ~$\bm{7.5}$ & ~$\bm{6.9}$ & $\bm{6.1}$ \\
   AC $\boldsymbol{L}_{-10,\delta}$ & $42.5$ & $13.2$ & ~$\bm{8.7}$ & ~$\bm{7.5}$ & ~$\bm{6.9}$ & $\bm{6.2}$\rule[-0.4em]{0em}{1em}  \\
   \hline
  \end{tabular}\\[1em]

  Wikipedia~\cite{wiki} 
($n=693$, $T=2$, $m=10$; AC: $\sigma=2$, $k=74$)
\begin{tabular}{|l c c c c c c|}
\hline \hline
   known labels\rule[0em]{0em}{1em} & $1\%$ & $5\%$ & $10\%$ & $15\%$ & $20\%$ & $25\%$\\ \hline
   SSL $\boldsymbol{L}_1$\rule[0em]{0em}{1em} & $68.2$ & $61.1$ & $53.6$ & $48.3$ & $44.1$ & $42.3$ \\
   SSL $\boldsymbol{L}_{-1,\delta}$ & $59.1$ & $52.3$ & $40.2$ & $36.3$ & $35.1$ & $34.1$ \\
   SSL $\boldsymbol{L}_{-10,\delta}$\rule[-0.4em]{0em}{1em} & $66.9$ & $57.2$ & $43.2$ & $38.7$ & $36.3$ & $34.9$ \\
\hline
   AC $\boldsymbol{L}_1$\rule[0em]{0em}{1em} & $\bm{48.4}$ & $\bm{39.4}$ & $\bm{31.4}$ & $\bm{30.1}$ & $\bm{28.7}$ & $\bm{27.9}$ \\
   AC $\boldsymbol{L}_{-1,\delta}$ & $\bm{48.4}$ & $\bm{39.4}$ & $\bm{31.4}$ & $\bm{30.0}$ & $\bm{28.7}$ & $\bm{27.9}$ \\
   AC $\boldsymbol{L}_{-10,\delta}$ & $\bm{48.4}$ & $\bm{39.4}$ & $\bm{31.4}$ & $\bm{30.1}$ & $\bm{28.8}$ & $\bm{27.9}$\rule[-0.4em]{0em}{1em}  \\
   \hline
  \end{tabular}\\[1em]

  Citeseer~\cite{lu2003link} 
($n=3\,312$, $T=1$, $m=6$; AC: $\sigma=2$, $k=130$)
  \begin{tabular}{|l c c c c c c|}
  \hline \hline
   known labels\rule[0em]{0em}{1em} & $1\%$ & $5\%$ & $10\%$ & $15\%$ & $20\%$ & $25\%$\\ \hline
   SSL $\boldsymbol{L}_1$\rule[0em]{0em}{1em} & $56.3$ & $44.1$ & $41.2$ & $38.5$ & $36.1$ & $34.7$ \\
   SSL $\boldsymbol{L}_{-1,\delta}$  & $\bm{52.4}$ & $39.0$ & $35.6$ & $32.6$ & $30.9$ & $\bm{29.5}$ \\
   SSL $\boldsymbol{L}_{-10,\delta}$\rule[-0.4em]{0em}{1em} & $68.6$ & $54.6$ & $48.5$ & $43.0$ & $39.7$ & $37.2$ \\
\hline
   AC $\boldsymbol{L}_1$\rule[0em]{0em}{1em} & $57.9$ & $\bm{34.3}$ & $\bm{32.1}$ & $\bm{31.3}$ & $\bm{30.0}$ & $\bm{29.4}$ \\
   AC $\boldsymbol{L}_{-1,\delta}$ & $57.9$ & $\bm{34.3}$ & $\bm{32.1}$ & $\bm{31.3}$ & $\bm{30.0}$ & $\bm{29.4}$ \\
   AC $\boldsymbol{L}_{-10,\delta}$ & $57.9$ & $\bm{34.3}$ & $\bm{32.1}$ & $\bm{31.3}$ & $\bm{30.0}$ & $\bm{29.4}$\rule[-0.4em]{0em}{1em} \\
   \hline
   \end{tabular} \\[1em]
   WebKB~\cite{craven1998learning} 
($n=187$, $T=1$, $m=5$; AC: $\sigma=2$, $k=15$)
   \begin{tabular}{|l c c c c c c|}
   \hline \hline
   known labels\rule[0em]{0em}{1em} & $1\%$ & $5\%$ & $10\%$ & $15\%$ & $20\%$ & $25\%$\\ \hline
   SSL $\boldsymbol{L}_1$\rule[0em]{0em}{1em} & $58.5$ & $49.0$ & $44.8$ & $44.3$ & $44.5$ & $44.4$ \\
   SSL $\boldsymbol{L}_{-1,\delta}$  & $49.9$ & $45.5$ & $40.7$ & $39.5$ & $39.9$ & $40.3$ \\
   SSL $\boldsymbol{L}_{-10,\delta}$\rule[-0.4em]{0em}{1em} & $52.3$ & $41.9$ & $38.0$ & $38.1$ & $36.8$ & $39.5$ \\
\hline
   AC $\boldsymbol{L}_1$\rule[0em]{0em}{1em} & $\bm{43.7}$ & $\bm{33.2}$ & $\bm{23.3}$ & $\bm{19.6}$ & $\bm{15.5}$ & $\bm{14.8}$ \\
   AC $\boldsymbol{L}_{-1,\delta}$ & $\bm{43.7}$ & $\bm{33.2}$ & $\bm{23.3}$ & $\bm{19.6}$ & $\bm{15.5}$ & $\bm{14.8}$ \\
   AC $\boldsymbol{L}_{-10,\delta}$ & $\bm{43.7}$ & $\bm{33.2}$ & $\bm{23.3}$ & $\bm{19.6}$ & $\bm{15.5}$ & $\bm{14.8}$\rule[-0.4em]{0em}{1em}  \\
   \hline
   \end{tabular} \\[1em]
  \end{center}
   \end{tiny}
  \end{minipage}
\vspace*{0.5em}
 \caption{Mean misclassification rate in percent of the multiclass Allen--Cahn scheme (\Cref{alg::multilayer_graph_allen_cahn_multiclass} denoted by ``AC'') using power mean Laplacians with different parameters $p$ in comparison with results from \cite{mercado2019generalized} (denoted by ``SSL'') using the same power mean Laplacians. For \Cref{alg::multilayer_graph_allen_cahn_multiclass}, the parameters $\epsilon=5 \cdot 10^{-3}$, $\omega_0=1\,000$, $c=\omega_0+3/\epsilon$, $\Delta t=0.01$, $\mathtt{tolerance}=10^{-6}$, and $\mathtt{max\_iter}=300$ are used. }\label{tab:results_pedro}
 \end{table}

We employ the multiclass multilayer Allen--Cahn classification scheme (\Cref{alg::multilayer_graph_allen_cahn_multiclass} from \Cref{sec:graph_allen_cahn_multilayer}) on those data sets using the power mean Laplacians $\boldsymbol{L}_1$, $\boldsymbol{L}_{-1,\delta}$, and $\boldsymbol{L}_{-10,\delta}$, where we set the parameters $\epsilon=5 \cdot 10^{-3}$, $\omega_0=1\,000$, $c=\omega_0+3/\epsilon$, $\Delta t=0.01$, $\mathtt{tolerance}=10^{-6}$, and $\mathtt{max\_iter}=300$. The resulting mean misclassification rates are shown in \Cref{tab:results_pedro} with the prefix ``AC''. We use the Gaussian kernel for the weight matricies~$\boldsymbol{W}$ with scaling parameter $\sigma$ as mentioned in the table for each data set. 

In general, we observe that \Cref{alg::multilayer_graph_allen_cahn_multiclass} has a lower misclassification rate than the SSL classification scheme~\cite{mercado2019generalized} for a ratio of known labels $\geq 5\%$, while the latter often performs better for $1\%$ known labels. Moreover, the value of~$p$ seems to have a higher influence on the misclassification rate for~\cite{mercado2019generalized}, whereas the influence of~$p$ is distinctly smaller in case of \Cref{alg::multilayer_graph_allen_cahn_multiclass}. Interestingly, the misclassification rates when applying the SSL classification scheme from~\cite{mercado2019generalized} also depend on~$p$ for the single layer data sets Citeseer, Cora, and WebKB, which is not the case for \Cref{alg::multilayer_graph_allen_cahn_multiclass}.
In addition, the results for the method~\cite{mercado2019generalized} are better in case of the Cora data set, whereas the results for \Cref{alg::multilayer_graph_allen_cahn_multiclass} in case of the WebKB data set are distinctly improved.

One likely explanation for the different behavior of the two considered methods is as follows.
Since the modified Ginzburg--Landau energy functional~$\tilde{E}$ in \eqref{equ:tilde_E_U} can be viewed as the loss function of our approach, one main difference between method~\cite{mercado2019generalized} and \Cref{alg::multilayer_graph_allen_cahn_multiclass} is the additional potential term $\frac{1}{2 \epsilon} \sum_{i=1}^{n} \left(\prod_{l=1}^{m} \frac{1}{4} \| \boldsymbol{u}_{i} - \boldsymbol{e}_{l} \|^{2}_{\ell^1} \right)$ in \eqref{equ:tilde_E_U}, which promotes more distinct class affiliations. 

\begin{table}
\begin{minipage}{0.49\textwidth}
\begin{tiny}
\begin{center}
BBC
\begin{tabular}{|l c c c c c c|}
\hline \hline
   known labels\rule[0em]{0em}{1em} & $1\%$ & $5\%$ & $10\%$ & $15\%$ & $20\%$ & $25\%$\\ \hline
   SSL $\boldsymbol{L}_1$\rule[0em]{0em}{1em} & 0.28 & 0.27 & 0.27 & 0.28 & 0.30 & 0.29 \\
   SSL $\boldsymbol{L}_{-1,\delta}$ & 0.30 & 0.30 & 0.32 & 0.31 & 0.33 & 0.31 \\
   SSL $\boldsymbol{L}_{-10,\delta}$\rule[-0.4em]{0em}{1em} & 0.32 & 0.34 & 0.34 & 0.34 & 0.34 & 0.34 \\
\hline
   AC $\boldsymbol{L}_1$\rule[0em]{0em}{1em} & 0.30 &  0.27 & 0.21 & 0.18 & 0.17 & 0.16 \\
   AC $\boldsymbol{L}_{-1,\delta}$ & 0.39 & 0.33 & 0.27 & 0.24 & 0.23 & 0.22 \\
   AC $\boldsymbol{L}_{-10,\delta}$\rule[-0.4em]{0em}{1em} & 0.78 & 0.73 & 0.65 & 0.63 & 0.63 & 0.61  \\
   \hline
  \end{tabular}
\end{center}
\end{tiny}
\end{minipage}
\hfill
\begin{minipage}{0.49\textwidth}
\begin{tiny}
\begin{center}
UCI
\begin{tabular}{|l c c c c c c|}
\hline \hline
   known labels\rule[0em]{0em}{1em} & $1\%$ & $5\%$ & $10\%$ & $15\%$ & $20\%$ & $25\%$\\ \hline
   SSL $\boldsymbol{L}_1$\rule[0em]{0em}{1em} & 11.2 & 11.5 & 11.3 & 11.4 & 11.5 & 11.3 \\
   SSL $\boldsymbol{L}_{-1,\delta}$ & 11.8 & 11.4 & 11.6 & 11.4 & 11.2 & 11.5 \\
   SSL $\boldsymbol{L}_{-10,\delta}$\rule[-0.4em]{0em}{1em} & 11.2 & 11.2 & 11.2 & 11.2 & 11.4 & 11.3 \\
\hline
   AC $\boldsymbol{L}_1$\rule[0em]{0em}{1em} & 3.26 & 3.16 & 3.03 & 3.08 & 2.83 & 2.78 \\
   AC $\boldsymbol{L}_{-1,\delta}$ & 4.57 & 4.49 & 4.21 & 4.17 & 4.21 & 3.96 \\
   AC $\boldsymbol{L}_{-10,\delta}$\rule[-0.4em]{0em}{1em} & 14.1 & 14.0 & 14.0 & 13.8 & 13.7 & 13.6 \\
   \hline
  \end{tabular}
\end{center}
\end{tiny}
\end{minipage}
\vspace*{0.5em}
\caption{Mean runtimes (over 10 runs) in seconds for BBC and UCI data set from \Cref{tab:results_pedro}.}
\label{tab:results_pedro_runtimes}
\end{table}

The observed runtimes behave differently depending on the number of nodes. For instance, for the BBC data set, the SSL classification scheme~\cite{mercado2019generalized} requires approx.\ $0.3$ seconds (average over 10 test runs) for each considered $p$ and known label ratio, while \Cref{alg::multilayer_graph_allen_cahn_multiclass} requires between 0.16 seconds and 0.78 seconds, cf.\ \Cref{tab:results_pedro_runtimes}.
Moreover, for the UCI data set, the SSL classification scheme~\cite{mercado2019generalized} has a runtime between 11.2 and 11.8 seconds, whereas \Cref{alg::multilayer_graph_allen_cahn_multiclass} takes between approx.\ $2.8$ and 14.1 seconds.

\subsection{Image data}
\label{sec:numerics:image}

Image segmentation tasks have been a key application for machine learning techniques for many years. While many methods rely on convolutions, image data can also be represented as a graph. A typical approach is to interpret each pixel of an image as a graph node, which is represented by its $3$-dimensional $8$-bit color channel (i.e.\ RGB) values, that range in the interval $[0, 255]\cap \mathbb{N}_0$.

One possible way to form a dense weight matrix $\boldsymbol{W}$ which also takes the spatial relation between the pixels into account is adding each pixel's location in the horizontal direction $x$ as well as in the vertical direction $y$ to the feature space, cf.~\cite{tomasi1998bilateral} for a related discussion in the computer vision context. This way, we can again operate on a fully connected graph with feature space dimension $d=5$. 

We employ our feature grouping approach from \Cref{sec:feature_grouping} and divide the $5$-dimensional feature space into two separate graph layers of a multilayer graph with $\mathcal{G}^{(1)}$ containing the $3$-dimensional color information and $\mathcal{G}^{(2)}$ the $2$-dimensional spatial information of the pixel locations. The information of the two layers is then recombined using the power mean Laplacian introduced in \eqref{equ:power_mean_laplacian}. This feature grouping enables the fast computation of the first eigenpairs of $\boldsymbol{L}_1$ belonging to the smallest eigenvalues by applying the NFFT-based fast summation to both low-dimensional graph layers. Furthermore, numerical experiments revealed, that this grouping of ``similar'' features achieves better image segmentation results than the single layer graph Laplacian on the full $5$-dimensional feature space. 

We test that approach on the image of \Cref{fig:beach_image} in \Cref{sec:graph_allen_cahn_multiclass}, which has about $9.7$ megapixels.
We vectorize the image data which leads to two data matrices $\boldsymbol{X}^{(1)} \in \mathbb{R}^{9\,734\,400 \times 3}$ and $\boldsymbol{X}^{(2)} \in \mathbb{R}^{9\,734\,400 \times 2}$, which are centered and then scaled to the boxes $[-1,1]^3$ and $[-1,1]^2$, respectively. We apply a multiclass approach with $m=4$ classes in order to segment the image into the four regions ``tree'', ``beach'', ``sea'' and ``sky''. The right image in \Cref{fig:beach_image} marks the classified pixels (approx.\ $4\,\%$), which are read out for the initialization of the Allen--Cahn multiclass method in order to act as the known label matrix $\boldsymbol{F}$. The rows of the initial solution matrix $\boldsymbol{U}_{0}$ are again initialized with the corresponding unit vector $\boldsymbol{e}_{j}^\top$ where available and $\frac{1}{m} \boldsymbol{1}^\top$ otherwise. For this example we use the scaling parameters $\sigma^{(1)}=1$ and $\sigma^{(2)}=4$ in the Gaussian kernel 
as well as compute $k=12$ eigenpairs.
We set the Allen--Cahn parameters $\epsilon=0.005$, $\omega_{0}=1\,000$, $c=\omega_{0}+3/\epsilon$, $\Delta t=0.01$, $\verb|max_iter|=500$ and $\verb|tolerance|=10^{-6}$.
Moreover, for the NFFT-based fast summation, we choose the 
NFFT parameters bandwidth $N=64$, window cutoff parameter $\texttt{m}_\text{NFFT}=5$, regularization length $\varepsilon_\mathrm{B}=1/16$, and regularization degree $\texttt{p}_\text{NFFT}=5$.

For $\boldsymbol{L}_1$, i.e., the case $p=1$, the computation of the $k=12$ eigenpairs using the Lanczos method and \Cref{alg::nfft_fastsum} requires approx.\ $980$ seconds. The Allen--Cahn scheme (\Cref{alg::multilayer_graph_allen_cahn_multiclass}) reaches its tolerance and terminates after $260$ iterations, which require approx.\ $950$ seconds. Note that all runtimes scale almost linearly with respect to the number of pixels~$n$, which makes the segmentation of larger images possible.

\begin{figure}
\subfloat[tree]{
{\setlength{\fboxsep}{0pt}\fbox{\includegraphics[width=.48\textwidth]{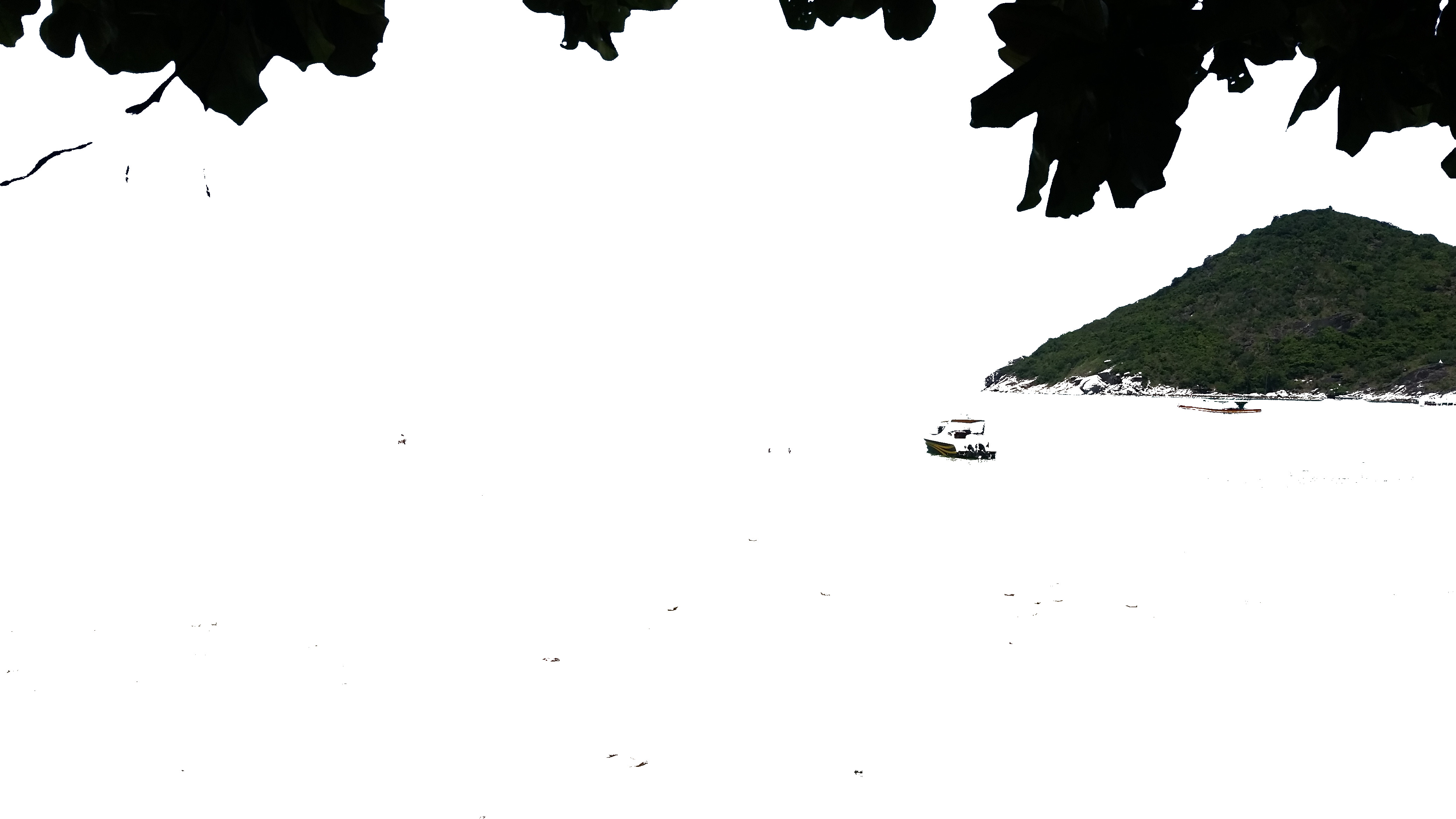}}}
}
\hfill
\subfloat[beach]{
{\setlength{\fboxsep}{0pt}\fbox{\includegraphics[width=.48\textwidth]{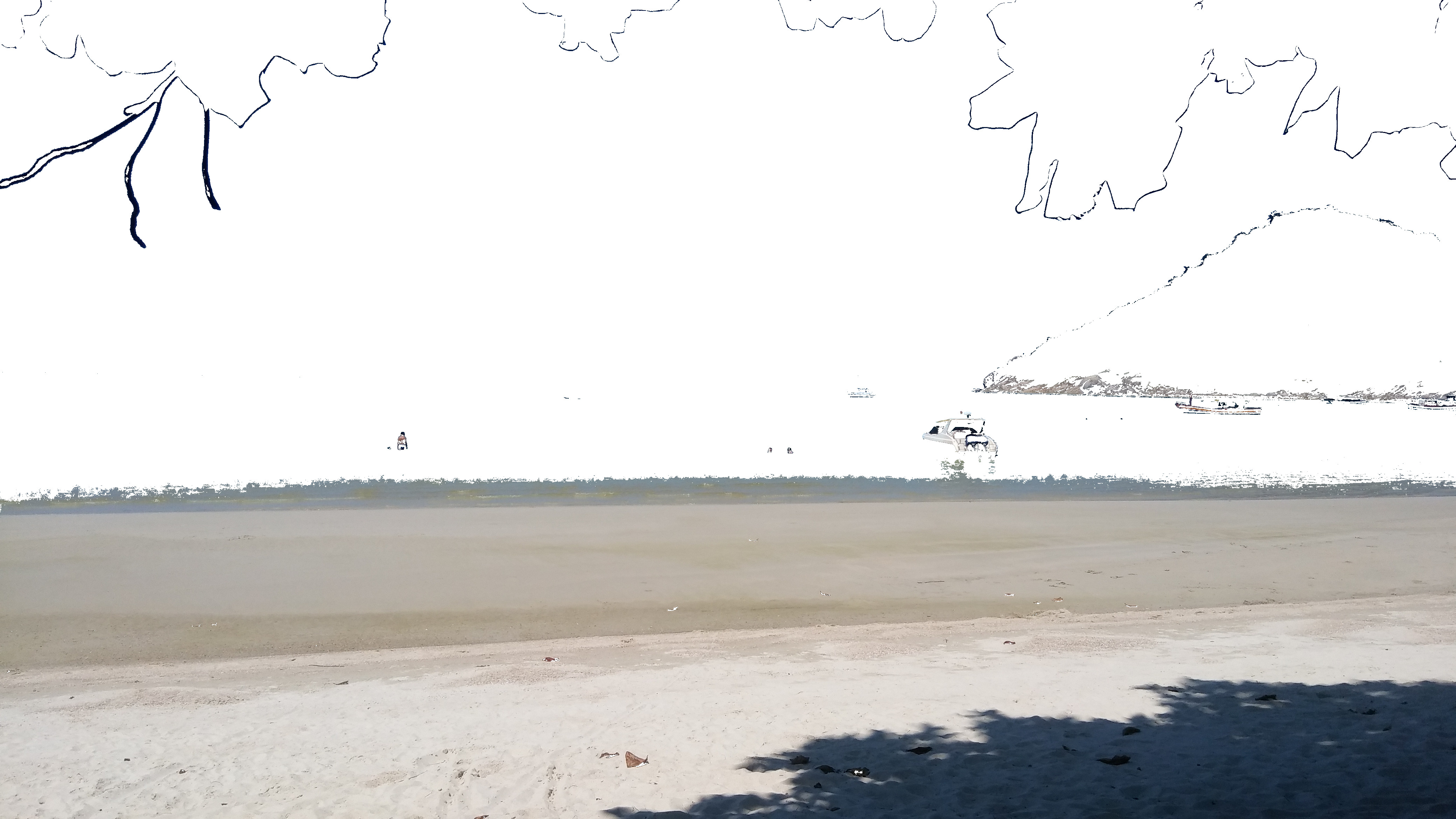}}}
}
\\
\subfloat[sea]{
{\setlength{\fboxsep}{0pt}\fbox{\includegraphics[width=.48\textwidth]{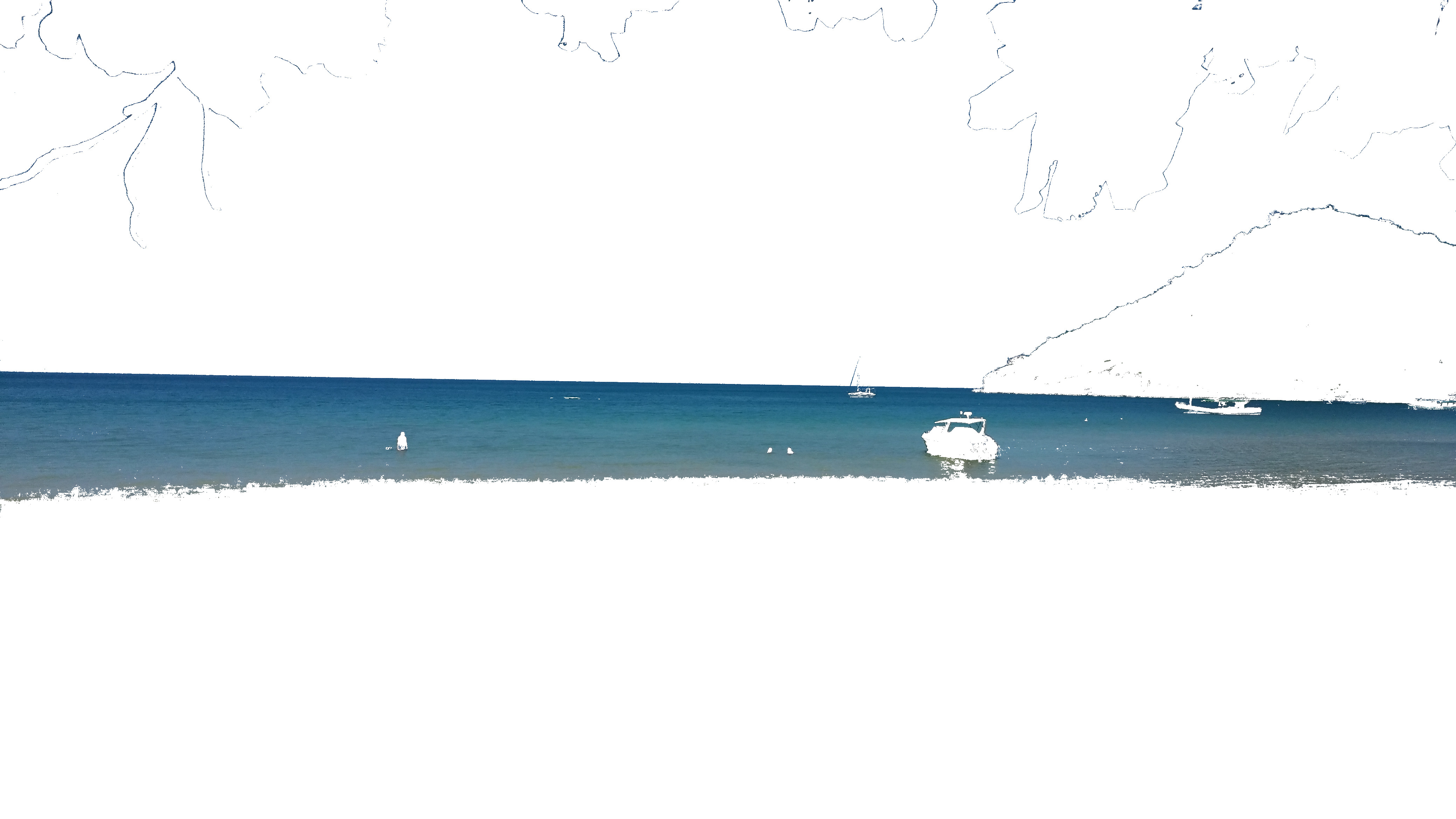}}}
}
\hfill
\subfloat[sky]{
{\setlength{\fboxsep}{0pt}\fbox{\includegraphics[width=.48\textwidth]{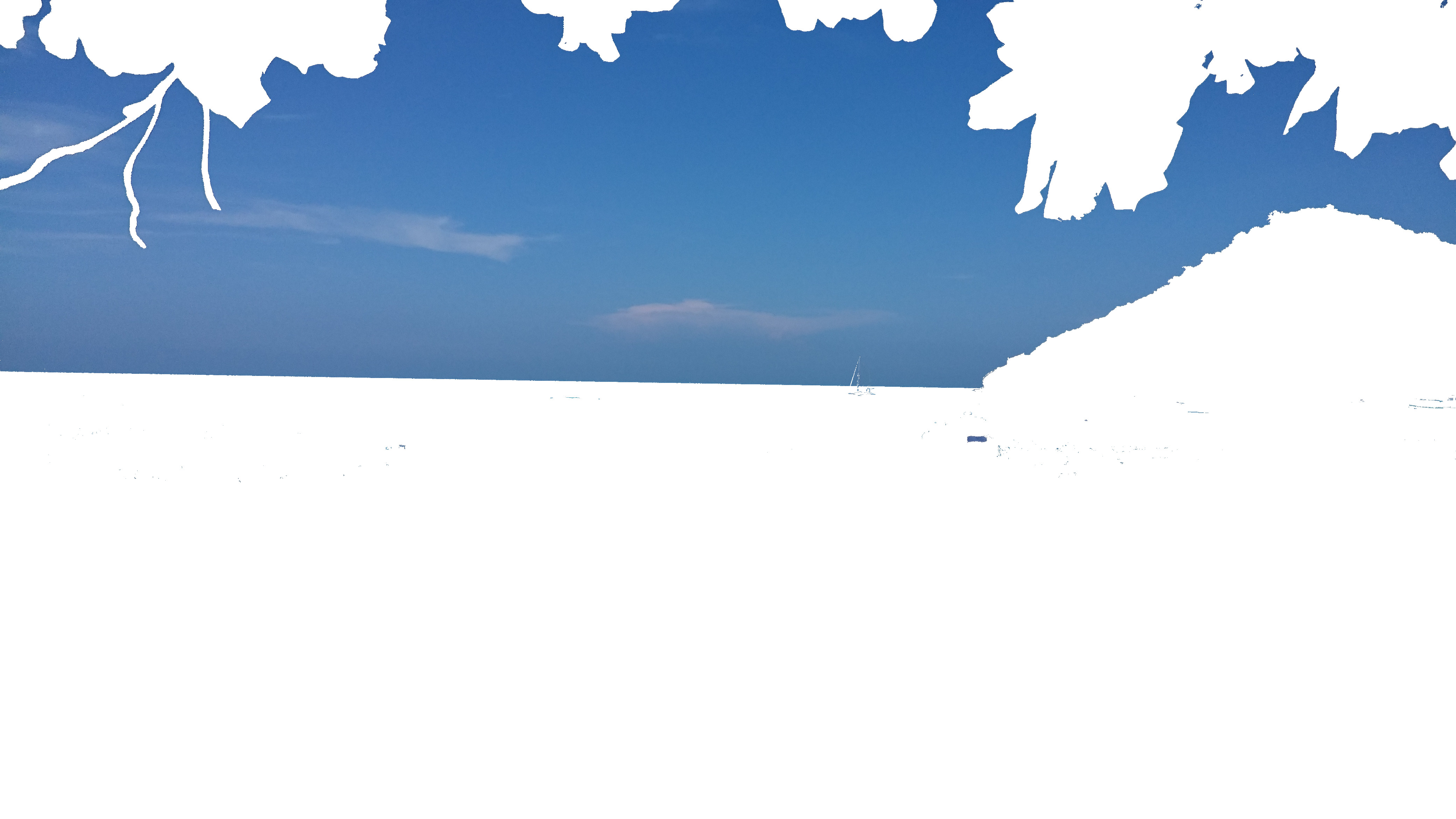}}}
}
\caption{$4$-class image segmentation result for test image from \Cref{fig:beach_image} using the $2$-layer power mean Laplacian~$\boldsymbol{L}_1$.}
 \label{fig:beach_image_classified}
\end{figure}

We then take the row-wise maximum of the output matrix $\boldsymbol{U}$ in order to make our prediction, to which class each pixel most likely belongs. \Cref{fig:beach_image_classified} shows the original pixel color for pixels, that are identified as belonging to the respective class and white pixels otherwise. 
Apart from minor systematic misclassifications for objects on the sea which do not possess an own class, the method segments the image very well.
When we consider $\boldsymbol{L}_{p,\delta}$ with $p=-10$ instead of $\boldsymbol{L}_1$, the classification results improve slightly, but the runtime for computing the eigenpairs increases to approx.\ $7090$ seconds since we have to use the PKSM in addition. \Cref{alg::multilayer_graph_allen_cahn_multiclass} requires again approx.\ $950$ seconds.

\begin{figure}
\subfloat[tree]{
{\setlength{\fboxsep}{0pt}\fbox{\includegraphics[width=.48\textwidth]{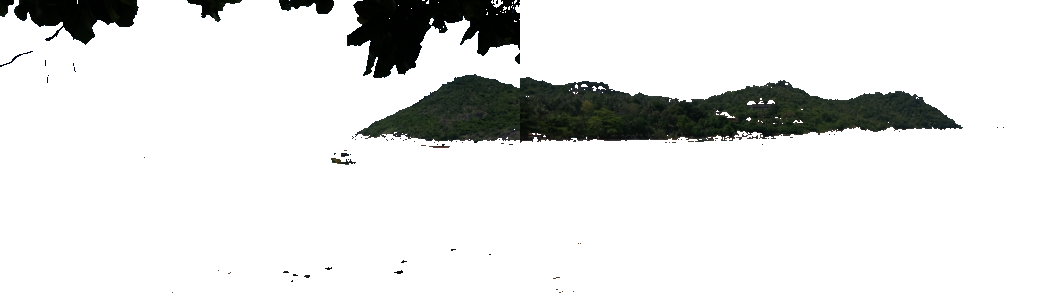}}}
}
\hfill
\subfloat[beach]{
{\setlength{\fboxsep}{0pt}\fbox{\includegraphics[width=.48\textwidth]{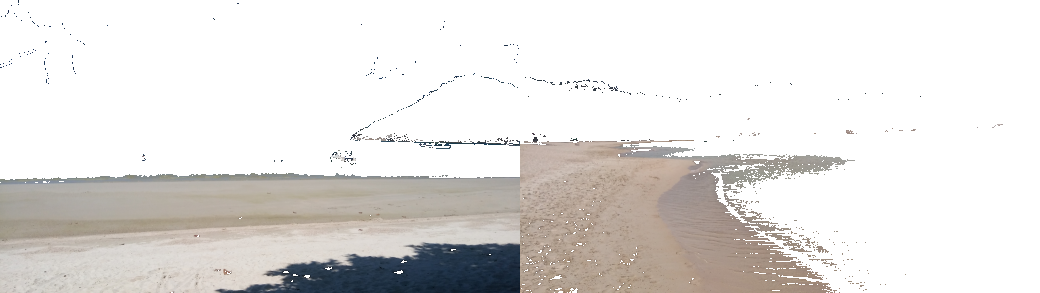}}}
}
\\
\subfloat[sea]{\label{fig:two_beach_images_transfer_labels_p1_sep_k35_maxit500:sea}
{\setlength{\fboxsep}{0pt}\fbox{\includegraphics[width=.48\textwidth]{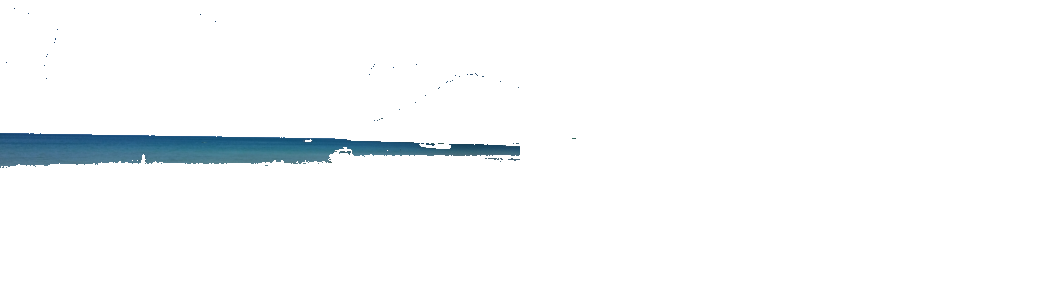}}}
}
\hfill
\subfloat[sky]{\label{fig:two_beach_images_transfer_labels_p1_sep_k35_maxit500:sky}
{\setlength{\fboxsep}{0pt}\fbox{\includegraphics[width=.48\textwidth]{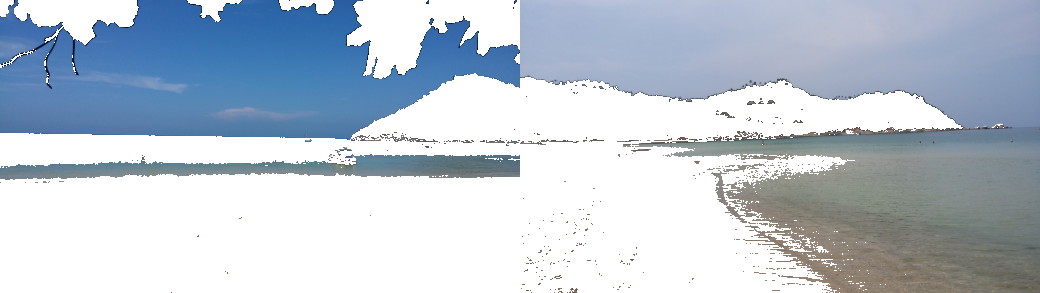}}}
}
\caption{$4$-class image segmentation result for label transfer to a second image using the 2-layer using the $2$-layer power mean Laplacian~$\boldsymbol{L}_1$.}
 \label{fig:two_beach_images_transfer_labels_p1_sep_k35_maxit500}
\end{figure}

In a second step, we consider a downscaled version of the same image as well as a similar image of the same size and concatenate both images to one image with $n=304\,720$ pixels. By keeping only the pre-labeled nodes from the first image we aim to transfer the known-label information not only to the image itself but to an unseen image consisting of the same classes. We use the same feature grouping and power mean Laplacian $\boldsymbol{L}_1$.
In the second graph layer~$\mathcal{G}^{(2)}$ containing the $2$-dimensional spatial information of the pixel locations, we now use two connected components, one for each partial image, and the corresponding graph Laplacian~$\boldsymbol{L}_{\mathrm{sym}}^{(2)}$ is block diagonal with two large densely populated blocks belonging to each partial image. For the Gaussian kernel, the NFFT-based fast summation, and the Allen--Cahn scheme, we use the same parameters as for the single image. Moreover, we increase the number of eigenpairs to $k=35$. Now, we only have around $2\%$ of pre-labeled nodes and obtain the classification results displayed in \Cref{fig:two_beach_images_transfer_labels_p1_sep_k35_maxit500} showing obvious misclassifications in \Cref{fig:two_beach_images_transfer_labels_p1_sep_k35_maxit500:sea} and~\ref{fig:two_beach_images_transfer_labels_p1_sep_k35_maxit500:sky}.
The computation of the $k=35$ eigenpairs requires approx.\ $3\,870$ seconds and the Allen--Cahn scheme additionally approx.\ $74$ seconds. The runtime for the eigenpairs computation can be drastically reduced by requesting fewer values, e.g., to approx.\ $55$ seconds for $k=27$ while achieving visually similar segmentation results as in \Cref{fig:two_beach_images_transfer_labels_p1_sep_k35_maxit500}. The reason for the larger runtime in case of $k=35$ eigenpairs is that the used Krylov-Schur method almost stagnates at $27$ eigenpairs.

\begin{figure}
\subfloat[tree]{
{\setlength{\fboxsep}{0pt}\fbox{\includegraphics[width=.48\textwidth]{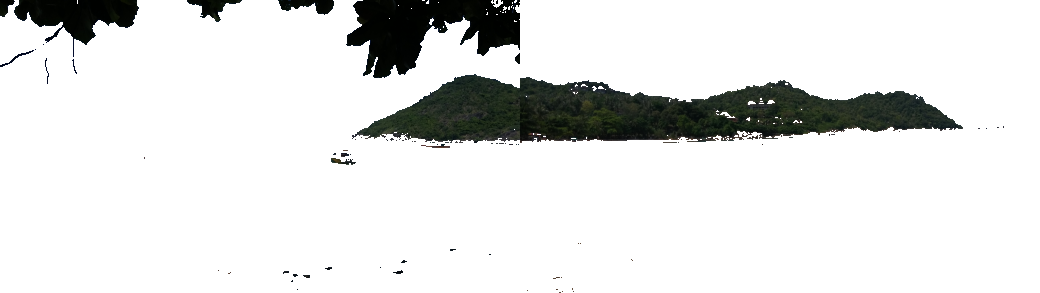}}}
}
\hfill
\subfloat[beach]{
{\setlength{\fboxsep}{0pt}\fbox{\includegraphics[width=.48\textwidth]{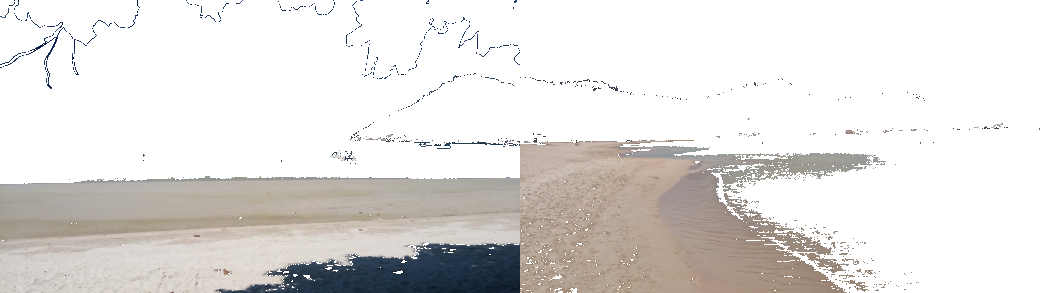}}}
}
\\
\subfloat[sea]{\label{fig:two_beach_images_transfer_labels_p-10_sep_k35_maxit500:sea}
{\setlength{\fboxsep}{0pt}\fbox{\includegraphics[width=.48\textwidth]{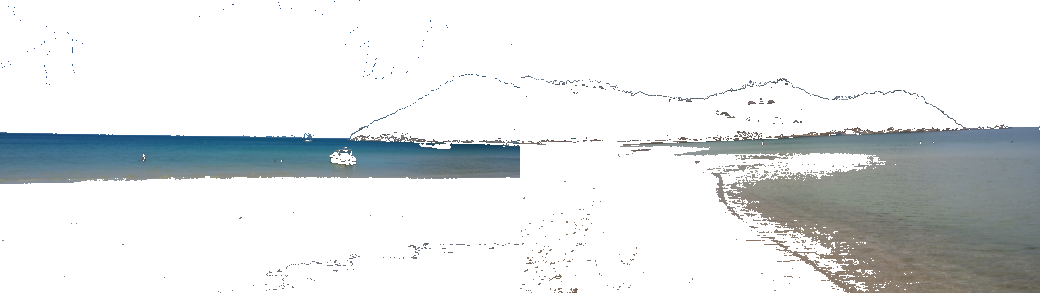}}}
}
\hfill
\subfloat[sky]{\label{fig:two_beach_images_transfer_labels_p-10_sep_k35_maxit500:sky}
{\setlength{\fboxsep}{0pt}\fbox{\includegraphics[width=.48\textwidth]{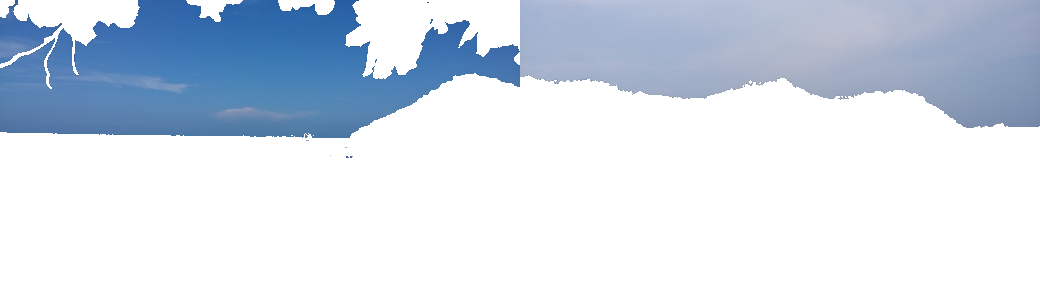}}}
}
\caption{$4$-class image segmentation result for label transfer to a second image using the 2-layer using the $2$-layer power mean Laplacian~$\boldsymbol{L}_{-10,\delta}$.}
 \label{fig:two_beach_images_transfer_labels_neg_p_sep_k35_maxit500}
\end{figure}

Similar to the observations made in \Cref{sec:numerics:SBM}, the result can be enhanced using a negative power $p<0$ in the power mean Laplacian. Here in our case, using $p=-10$ leads to much better results as displayed in \Cref{fig:two_beach_images_transfer_labels_neg_p_sep_k35_maxit500}. The correct detection of the different areas in the second image is a strong result given the varying colors in the two images, especially in the sea and sky regions in \Cref{fig:two_beach_images_transfer_labels_p-10_sep_k35_maxit500:sea} and~\ref{fig:two_beach_images_transfer_labels_p-10_sep_k35_maxit500:sky}, respectively.
The computation of the $k=35$ eigenpairs requires approx.\ $585$ seconds and the Allen--Cahn scheme approx.\ $58$ seconds. Reducing $k$ to $30$ yields a further reduction of the computation time of the eigenpairs to approx.\ $438$ seconds while still achieving similar results as in \Cref{fig:two_beach_images_transfer_labels_neg_p_sep_k35_maxit500}. Setting $k<30$ worsens the segmentation results similar to the ones in \Cref{fig:two_beach_images_transfer_labels_p1_sep_k35_maxit500}. We discuss the influence of the parameter $k$ in more detail in the following subsection.

\subsection{Hyperspectral data}\label{sec:numerics:hyperspectral}

Finally, we tackle a problem that necessitates the full arsenal of methods derived in this paper. We consider the urban mapping problem posed by the Pavia center data set \cite{plaza2009recent} as an example for the classification of the vast amounts of earth observation data gathered these days.

The original image size of 1096 by 1096 pixels contains valid ground truth labels for $n=148\,152$ of those pixels. The considered graph contains these labeled nodes. This together with the considerable number of classes $m=9$ as well as the high feature space dimension arising from $102$ atmospherically corrected hyperspectral frequency bands make this classification problem a demanding task. The frequently discussed combination of spectral and spatial information in the context of hyperspectral image classification \cite{plaza2009recent,wang2016spectral,fang2018semi} is here accomplished by adding the $x$- and $y$-pixel coordinates to the feature space, leading to a total of $d=104$ feature variables. 

As, depending on the assembling strategy of the weight matrix, the memory requirement and/or the runtime of forming the Laplacian matrices would become infeasible, we enable the efficient numerical treatment of the problem by \Cref{alg::multilayer_graph_allen_cahn_multiclass} using our feature grouping approach from \Cref{sec:feature_grouping} for the $104$-dimensional feature space and the NFFT-based fast summation as discussed in \Cref{sec:nfft_fastsum} and given in \Cref{alg::nfft_fastsum}. Note, that we refrain from data preprocessing beyond grouping the features into layers and directly work on the raw hyperspectral data. 
For the Gaussian kernel, we set the scaling parameter to $\sigma=8000$ for features involving hyperspectral frequency bands, which take values between 0 and 8000, and $\sigma=2\cdot 1095$ for coordinates layers, which have pixel coordinates between 1 and 1096.
We set the Allen--Cahn parameters $\epsilon=0.5$, $\omega_0=10\,000$, $c=\omega_0+3/\epsilon$, $\Delta t=0.01$, $\verb|max_iter|=300$, $\verb|tolerance|=10^{-6}$.
For the NFFT-based fast summation, we choose the
NFFT parameters bandwidth $N=64$, window cutoff parameter $\texttt{m}_\text{NFFT}=3$, regularization length $\varepsilon_\mathrm{B}=1/16$, and regularization degree $\texttt{p}_\text{NFFT}=3$.
Moreover, we use 5 percent random known labels per class. 

\begin{table}
\begin{center}
\begin{small}
\begin{tabular}{|c l | c || c c c|}
\hline \hline
  layers & type & \#tests & $k=20$ & $k=40$ & $k=120$ \\
\hline
  1 & 2 bands (rand.) & 100 & $0.760\pm0.031$ & --- & --- \\
  2 & 2 bands (rand.) + coord. & 100 & $0.850\pm0.043$ & $0.859\pm0.062$ & --- \\
  2 & 2 bands (det.) + coord. & 51 & $0.864\pm0.036$ & $0.879\pm0.034$& --- \\
  52 & 51 $\times$ 2 bands (det.) + coord. & 100 & $0.928\pm0.001$ & $0.942\pm0.001$ & $\bm{0.957}\pm0.001$ \\
\hline
  1 & 3 bands (rand.) & 100 & $0.787\pm0.028$ & --- & --- \\
  2 & 3 bands (rand.) + coord. & 100 & $0.885\pm0.040$ & $0.896\pm0.035$ & --- \\
  2 & 3 bands (det.) + coord. & 34 & $0.915\pm0.020$ & $0.920\pm0.014$ & --- \\
  35 & 34 $\times$ 3 bands (det.) + coord. & 100 & $0.930\pm0.001$ & $0.943\pm0.001$ & $\bm{0.959}\pm0.001$ \\
\hline
\hline
\end{tabular}
\end{small}
\end{center}
\vspace{0.5em}
\caption{Average classification accuracies and standard deviations for Pavia center data set \cite{plaza2009recent} using \Cref{alg::multilayer_graph_allen_cahn_multiclass} in combination with \Cref{alg::nfft_fastsum} for 5\% known labels per class. In the multilayer cases, $p=-10$ is used. ``rand.'' means random and ``det.'' deterministic frequency band selection, ``coord.'' means coordinates layer. The Gaussian kernel with scaling parameter $\sigma=8000$ is used for frequency bands and $\sigma=2\cdot 1095$ for the coordinates layer.}
\label{table::Pavia_center:1}
\end{table}

\begin{sloppypar}
The results shown in \Cref{table::Pavia_center:1} illustrate the classification performance for different data modeling approaches. The lowest accuracies are obtained in the cases where we consider single hyperspectral layers of two and three bands, respectively. In each case, we draw 100 random combinations with replacement and average over the 100 test runs. Following the approach presented in \Cref{sec:numerics:image}, these results can be distinctly improved by adding the pixel coordinates as a second layer and using the power mean Laplacian with $p=-10$. Interestingly, a deterministic band selection improves the achieved accuracy further. Here, we compute the results averaged over the 51 combinations $(1,52)$, $(2,53)$, \ldots, $(51,102)$ of two bands and the 34 combinations $(1,35,69)$, $(2,36,70)$, \ldots, $(34,68,102)$ of three bands respectively. For all experiments with only one hyperspectral layer, we obtained a rather high variance in the classification accuracy. In addition to the random choice of hyperspectral bands, these deviations originate from the different random choices of pre-labeled nodes in each test run.
\end{sloppypar}

The best results, however, are obtained by utilizing all band information by employing our feature grouping approach from \Cref{sec:feature_grouping}. Here, we use all bands with two bands per layer ($(1,52)$, $(2,53)$, \ldots, $(51,102)$), resulting in 51 frequency bands and 1 coordinate layer, as well as all bands with three bands per layer ($(1,35,69)$, $(2,36,70)$, \ldots, $(34,68,102)$), resulting in 34+1 layers respectively. 

Within the range of the number $k$ of eigenpairs of the respective graph Laplacian we consider in our experiments, a higher number of eigenpairs tends to improve the achieved accuracy at the cost of an increased runtime. Depending of the feature space dimension, a saturation of the accuracy typically sets in at some point. Choosing $k$ too large, however, bears the risk of including noisy eigenvectors, resulting in worse accuracies or potential convergence issues for the eigenvector computations by the Lanczos algorithm. Similar observations can be made for different data sets.

Furthermore, the results can be distinctly improved in the 52 layers case of two bands per layer by modifying the scaling parameter $\sigma=8000/2$ for the frequency bands layers and $\sigma=1095/2$ for the coordinates layer.
We achieve an average accuracy of $0.958\pm0.001$ when using $k=40$ eigenvectors and of $0.975\pm0.001$ for $k=120$. We obtain almost the same numbers using the same scaling parameters in the case of all bands with three bands per hyperspectral layer and the additional coordinates layer. Moreover, varying the percentage of known labels per class has only a relatively small influence on the accuracies, cf.\ \Cref{table::Pavia_center:2}. In particular, for only 0.5\% known labels per class, we achieve average accuracies of $0.972\pm0.003$ and $0.974\pm0.003$ for two frequency bands per layer and three frequency bands per layer, respectively. Here, in the two frequency bands per layer case, the eigenpair computation takes approx.\ $12\,800$ seconds as well as approx.\ $26\,900$ seconds for the three bands per layer case. These increased runtimes in comparison to earlier examples are caused by the increased number of eigenvalues as well as graph layers. The latter, however, only enters the computational complexity linearly as discussed in \Cref{sec:feature_grouping}. In both cases, the Allen--Cahn scheme requires approx.\ $71$ seconds on average. Note, that the computation time for the eigenpairs could be easily reduced on a many-core computer using parallelization.\footnote{For instance, we tested the two frequency bands per layer case from \Cref{table::Pavia_center:2} using a rather simple parallelization based on the \texttt{spmd} statement of the \textsc{MATLAB} \textit{Parallel Computing Toolbox}. On a computer with \mbox{4 x} Intel Xeon E5-4640 each with 8 x 2.40~GHz cores (in total 32 cores / 64 threads), we observed a runtime of 849 seconds for the eigenpair computation using 52 \textsc{MATLAB} processes, including the setup time for these 52 processes. For comparison, the single threaded case required $15\,770$ seconds on the same computer, resulting in a speedup of approx.\ $18.6$ and a parallel efficiency of approx.\ $36\%$.
Moreover, for the three frequency bands per layer case, we observed a runtime of $3\,120$ seconds for the eigenpair computation using 35 processes as well as $54\,417$ seconds for the single threaded case, which yields a speedup of approx.\ $17.4$ and a parallel efficiency of approx.\ $50\%$.
Please note, that the runtimes in this footnote can not be directly compared to the previously stated runtimes as the measurements were performed on another machine with a different CPU type.
}

Similarly, applying the same approach to other hyperspectral data sets like the Pavia university data set, another data set presented in \cite{plaza2009recent}, or the Indian pines data set \cite{PURR1947} also yields mean classification accuracies above $0.97$ given $5\%$ and $10\%$ pre-known labels, respectively. 

\begin{table}
\begin{center}
\begin{small}
\begin{tabular}{|l |c c c c|}
\hline \hline
  & \multicolumn{4}{c|}{known labels per class} \\
  type & 0.25\% & 0.5\% & 1\% & 5\% \\
\hline
  51 $\times$ 2 bands (det.) + coord. & $0.965\pm0.005$ & $\bm{0.972}\pm0.003$ & $0.968\pm0.002$ & $0.975\pm0.001$ \\
\hline
  34 $\times$ 3 bands (det.) + coord. & $0.967\pm0.005$ & $\bm{0.974}\pm0.003$ & $0.972\pm0.002$ & $0.977\pm0.001$ \\
\hline
\hline
\end{tabular}
\end{small}
\end{center}
\vspace{0.5em}
\caption{Average accuracies and standard deviations over 100 test runs for Pavia center data set \cite{plaza2009recent} using \Cref{alg::multilayer_graph_allen_cahn_multiclass} with $p=-10$ and $k=120$ in combination with \Cref{alg::nfft_fastsum} for varying percentage of known labels per class. ``det.'' means deterministic frequency band selection and ``coord.'' means coordinates layer. The Gaussian kernel with scaling parameter $\sigma=8000/2$ is used for frequency bands and $\sigma=1095/2$ for the coordinates layer.}
\label{table::Pavia_center:2}
\end{table}

Note that there are many results for different classification approaches of the Pavia center data set available in the literature with varying foci (e.g.\ feature selection, maximizing classification accuracy, minimizing pre-labeled data) and classification accuracies. While there are examples for the application of ``classical'' methods \cite{plaza2009recent,wang2016data}, some authors reduce the feature space dimension by feature selection techniques \cite{dalla2011evolution,davari2017gmm}, whereas the top classification results are achieved with highly specialized convolutional neural network (CNN) architectures \cite{makantasis2015deep,lin2020spatial}.
Our results can compete with most results presented in the literature. Better results have been reported for problem-tailored CNN architectures and certain support vector machine classifiers \cite{fauvel2012advances,sun2012hyperspectral}.
However, we emphasize again that our method operates directly on the raw data without feature selection and data preprocessing beyond grouping the features into layers, and that it does not require excessive compute power or specialized hardware. In fact, all numerical experiments presented in this paper can be run on an average laptop computer.

\section{Conclusion}

We have studied the applicability of the power mean Laplacian in the context of diffuse interface-based semi-supervised learning for multilayer networks.
The presented feature grouping approach allowed us to tackle data sets with a larger dimension of the feature space than it was previously possible with standard NFFT-based fast matrix vector products.

\section*{Acknowledgement}
\begin{sloppypar}
T.\ Volkmer gratefully acknowledges partial funding by the S\"achsische Aufbaubank -- F\"orderbank -- (SAB) 100378180. The authors are indebted to the anonymous referees for their helpful comments.
\end{sloppypar}

\begin{appendix}

\section{Binary Allen--Cahn}\label{sec:binary_allen_cahn}
In $1958$, John W.\ Cahn and John E.\ Hilliard introduced a phase-field approach to describe phase separation phenomena in general solutions \cite{CahnHilliard}. The derivation of the partial differential equation, which is today called the Cahn--Hilliard equation, assumed mass conservation in the physical system.

In $1979$, after investigating metal alloys for several years, John W. Cahn and Sam Allen proposed a very similar approach. The Allen--Cahn equation \cite{AllenCahn} can be interpreted as a variant of the Cahn--Hilliard equation without the mass conservation condition.
In both partial differential equations, a scalar field $u\colon \Omega \rightarrow [-1,1] \subset \mathbb{R}$ describes the two components on a typically $3$-dimensional domain $\Omega$. The concentration of the components is then given by $\frac{1}{2}(u+1)$ and $\frac{1}{2}(1-u)$, respectively. Pure phases of the solution are represented by the values $1$ and $-1$, respectively, and a value of $0$ means a perfect mixture of both components.

In order to describe the evolution of the scalar field $u$ as a function of time, the Ginzburg--Landau energy functional is defined as in~\eqref{GinzburgLandau}.
The parameter $\epsilon$ weights the gradient term proportionally, which represents the Dirichlet energy that penalizes strong concentration gradients. As there naturally exist large concentration gradients at the phase boundaries, this gradient term enforces a rounded shape of the regions, as the ratio between circumference and area of a region is minimal for a circle. The interface parameter $\epsilon$ also weights the potential function $\psi (u)$ inversely proportionally. This $\psi (u)$ describes the chemical interaction energy at a single point $\boldsymbol{x}\in\Omega$ given its current concentration distribution $u(\boldsymbol{x})$. In order for the system to exhibit a phase separation behavior the potential function $\psi (u)$ must have energetic minima at or close to the pure phases $u=1$ and $u=-1$, respectively. While the original work~\cite{CahnHilliard} employs a logarithmic potential, in practice, often polynomial approximations such as 
\begin{equation}\label{defpsi}
\psi (u) = \frac{1}{4} (u^{2}-1)^{2},
\end{equation}
are used.
Standard solution methods for partial differential equations, such as the finite element method or the finite difference method, can be used to solve \eqref{AllenCahn}, given suitable initial conditions as well as boundary conditions. 
Here, the time derivative as well as the spatial derivative must be discretized. For the spatial discretization, the domain $\Omega$ must be triangularized, so that the equation is solved on a finite set of grid points instead of the continuous domain. 

Using the Allen--Cahn equation for a binary classification task can be motivated by identifying the grid points from the spatial discretization of the physical domain $\Omega$ with vertices $x_{i} \in \mathcal{V}$ of a graph~$\mathcal{G}$, cf.~\cite{BertozziFlenner2012DiffuseHighdimClassif}, which we identify with corresponding feature vectors. In doing so, one defines the quantity $u$ of the phase-field description on the finite set of vertices by replacing $u$ by a vector $\boldsymbol{u}\in\mathbb{R}^{n}$ with one entry per vertex~$x_i$. The goal in the binary classification case is to identify vertices with a corresponding entry in $\boldsymbol{u}$ close to $1$ with the first class and an entry close to $-1$ with the second class, belonging to either pure component in the phase-field formulation.

As the method is semi-supervised, we have a subset of pre-labeled graph vertices at our disposal. We encode this information in a vector $\boldsymbol{f} \in \mathbb{R}^n$ by setting $f_i=1$ if $x_i \in \mathcal{V}$ belongs to the first class, $f_i=-1$ if $x_i$ belongs to the second class, and $f_i=0$ otherwise.
In order to enforce the correct classification of this pre-labeled data, an additional term is included in the Ginzburg--Landau energy functional \eqref{GinzburgLandau}. Inspired by applications of the Cahn--Hilliard equation in image inpainting, \cite{BertozziFlenner2012DiffuseHighdimClassif} suggests a penalty term of the form $\frac{1}{2} \omega(x_i) (f_i-u_i)^{2}$ for a least squares fit to the data with $x_i \in \mathcal{V}$, where $\omega(x_i)$ is some (usually large) constant $\omega_{0}$ for pre-labeled vertices and $0$ for unlabeled vertices. 
This way, a deviation in $\boldsymbol{u}$ from $\boldsymbol{f}$ gets penalized in the energy functional that we seek to minimize. It can, however, still be energetically beneficial for $\boldsymbol{u}$ to deviate from $\boldsymbol{f}$ at some vertices for the sake of shorter interface lengths and thus a smaller Dirichlet energy. This way, the method even remains stable given few misclassified training samples. In the classification context, the interface represents the decision boundary. Thus, the minimization of interface lengths in the graph setting prevents labels to change across nodes in clustered regions with high node similarities, which are expressed by large edge weights. The choice of $\omega_0$ controls the trade-off between the classical Ginzburg--Landau energy and the least squares term. Choosing $\omega_0$ too small bears the risk of underfitting while choosing it too large may cause overfitting to the pre-classified data. With this modification, the discretized Ginzburg--Landau functional becomes
\begin{equation}\label{GinzburgLandaumodified}
\tilde{E}(\boldsymbol{u}) = \frac{\epsilon}{2} \boldsymbol{u}^\top\boldsymbol{L}_{\mathrm{sym}} \boldsymbol{u} + \frac{1}{\epsilon} \sum_{i=1}^{n} \frac{1}{4} ( u_{i}^{2} - 1)^{2} + \frac{1}{2} \sum_{i=1}^{n} \omega(x_{i})(f_{i}-u_{i})^{2},
\end{equation}
where the continuous integrals in \eqref{GinzburgLandau} become sums over the vertex set in the discrete graph setting and the potential $\psi$ from \eqref{defpsi} was inserted. This modified Ginzburg--Landau energy functional~$\tilde{E}$ forms the basis for the semi-supervised classification technique considered in this work as~$\tilde{E}$ can be viewed as the method's loss function. Note, that this loss function contains the potential function term $\frac{1}{\epsilon} \sum_{i=1}^n \psi (u_i)$ in addition to a regularized least squares fit approach for semi-supervised learning on graphs that is frequently used in the literature \cite{zhou2004learning,NFFTmeetsKrylov2018,mercado2019generalized}. 

In order to solve the Allen--Cahn equation \eqref{AllenCahn} for the modified Ginzburg--Landau energy functional \eqref{GinzburgLandaumodified}, this section presents a suitable convexity splitting approach for the binary classification setting. In this case, the numerical scheme treating the convex part of the Ginzburg--Landau functional implicitly and the concave part explicitly reads
\begin{equation}\label{convexitysplittingscheme}
\frac{\boldsymbol{u}^{l+1}-\boldsymbol{u}^{l}}{\Delta t} = -\frac{\partial E_1}{\partial u} (\boldsymbol{u}^{l+1}) + \frac{\partial E_2}{\partial u} (\boldsymbol{u}^{l})
\end{equation}
with iterates $\boldsymbol{u}^{l} \in \mathbb{R}^{n}$, time step size $\Delta t \in \mathbb{R}$ and index $l$ for the time step. 

A possible splitting for the modified Ginzburg--Landau functional~$\tilde{E}(\boldsymbol{u})$ in~\eqref{GinzburgLandaumodified} is presented in \cite{BertozziFlenner2012DiffuseHighdimClassif}, where a productive zero is inserted into $\tilde{E}(\boldsymbol{u})$ by adding and subtracting the convex function $\frac{c}{2} \boldsymbol{u}^\top \boldsymbol{u}$ with a constant $c>0$. The resulting functional is then split such that
\begin{equation}\label{E1}
\tilde{E}_1(\boldsymbol{u}) = \frac{\epsilon}{2} \boldsymbol{u}^\top \boldsymbol{L}_{\mathrm{sym}} \boldsymbol{u} + \frac{c}{2} \boldsymbol{u}^\top \boldsymbol{u}
\end{equation}
and
\begin{equation}\label{E2}
\tilde{E}_2(\boldsymbol{u}) = \frac{c}{2} \boldsymbol{u}^\top \boldsymbol{u} -\frac{1}{4 \epsilon} \sum_{i=1}^{n} ( u_{i}^{2} - 1)^{2} - \frac{1}{2} \sum_{i=1}^{n} \omega(x_{i})(f_{i}-u_{i})^{2}.
\end{equation}
While $\tilde{E}_1$ is strictly convex for all $\epsilon,c>0$, the parameters have to
be chosen
\begin{equation*}
c > \omega_{0} + \frac{3\overline{u}^2-1}{\epsilon}, \quad \overline{u} = \max_{i=1, \dots , n} | u_i |,
\end{equation*}
to ensure the strict concavity of $-\tilde{E}_2$, cf.~\cite{BertozziFlenner2012DiffuseHighdimClassif}.

Inserting $\tilde{E}_1$ and $\tilde{E}_2$ into the iteration scheme~\eqref{convexitysplittingscheme}
yields
\begin{equation*}
\frac{\boldsymbol{u}^{l+1}-\boldsymbol{u}^{l}}{\Delta t} = - \epsilon \boldsymbol{L}_{\mathrm{sym}} \boldsymbol{u}^{l+1} - c \boldsymbol{u}^{l+1} + c\boldsymbol{u}^{l} - \frac{1}{\epsilon} \left( (\boldsymbol{u}^{l})^{3} - \boldsymbol{u}^{l}\right) - \boldsymbol{\omega} \, (\boldsymbol{u}^{l}-\boldsymbol{f}),
\end{equation*}
where we set $\boldsymbol{\omega}:=\diag\left(\omega(x_{i})_{i=1}^n\right)$ and the power in $(\boldsymbol{u}^{l})^{3}$ is to be understood elementwise.
Now, the solution $\boldsymbol{u}$ is restricted to the ansatz $\boldsymbol{\Phi}_k \boldsymbol{v}$, where 
$\boldsymbol{\Phi}_k:=(\boldsymbol{\phi}_1,\ldots,\boldsymbol{\phi}_k)\in\mathbb{R}^{n\times k}$ is the matrix of the eigenvectors belonging to the $k$ smallest eigenvalues $\lambda_1,\ldots,\lambda_k$ of $\boldsymbol{L}_{\mathrm{sym}}$ and $\boldsymbol{v}\in\mathbb{R}^k$ is a coefficient vector.
Since $\boldsymbol{L}_{\mathrm{sym}}\boldsymbol{\Phi}_k = \boldsymbol{\Phi}_k \boldsymbol{\Lambda}_k$ with $\boldsymbol{\Lambda}_k = \diag(\lambda_{1},\lambda_{2},\dots ,\lambda_{k}) \in \mathbb{R}^{k \times k}$, we obtain
\begin{equation*}
\frac{\boldsymbol{\Phi}_k \boldsymbol{v}^{l+1}-\boldsymbol{\Phi}_k \boldsymbol{v}^{l}}{\Delta t} = - \epsilon \boldsymbol{\Phi}_k \boldsymbol{\Lambda}_k \boldsymbol{v}^{l+1} - c\boldsymbol{\Phi}_k \boldsymbol{v}^{l+1} + c\boldsymbol{\Phi}_k \boldsymbol{v}^{l} - \frac{1}{\epsilon} ((\boldsymbol{\Phi}_k \boldsymbol{v}^{l})^{3} - \boldsymbol{\Phi}_k \boldsymbol{v}^{l}) - \boldsymbol{\omega} \, (\boldsymbol{\Phi}_k \boldsymbol{v}^{l}-\boldsymbol{f}).
\end{equation*}
Multiplication from the left with $\boldsymbol{\Phi}_k^\top$, simplifications and rearranging finally yields the iteration rule
\begin{equation}\label{iterationfinal}
v_{r}^{l+1} = \frac{1}{1+\epsilon \lambda_{r} \, \Delta t + c \Delta t} \left[ \left(1+ c \, \Delta t + \frac{\Delta t}{\epsilon} \right) v_{r}^{l} - \frac{\Delta t}{\epsilon} b_{r}^{l} - \Delta t \, d_{r}^{l} \right],
\; r=1,\ldots,k,
\end{equation}
with
$\boldsymbol{b}^{l} = \epsilon^{-1} \boldsymbol{\Phi}_k^\top (\boldsymbol{\Phi}_k \boldsymbol{v}^{l})^{3} \in \mathbb{R}^{k}$
and
$\boldsymbol{d}^{l} = \boldsymbol{\Phi}_k^\top \boldsymbol{\omega} (\boldsymbol{\Phi}_k \boldsymbol{v}^{l}-\boldsymbol{f}) \in \mathbb{R}^{k}$.

\section{Numerical solution of the multiclass graph Allen--Cahn scheme}\label{sec:convexitysplittingmulticlass}

\begin{sloppypar}
	In \Cref{sec:graph_allen_cahn_multiclass}, we present the multiclass graph Allen--Cahn classifier and derive a convexity splitting scheme minimizing the modified Ginzburg--Landau energy functional \eqref{equ:tilde_E_U}. We now provide details on the efficient numerical solution of the numerical scheme \eqref{equ:U_lp1}.
	
	First, inserting the eigendecomposition $\boldsymbol{L}_{\mathrm{sym}} = \boldsymbol{\Phi} \boldsymbol{\Lambda} \boldsymbol{\Phi}^\top$ into \eqref{equ:U_lp1} gives
	\begin{equation}\label{equ:inv_B_multiclass}
	\boldsymbol{B}^{-1} = \big[(1+c(\Delta t)) \boldsymbol{\Phi} \boldsymbol{\Phi}^\top + \epsilon (\Delta t) \boldsymbol{\Phi} \boldsymbol{\Lambda} \boldsymbol{\Phi}^\top \big]^{-1} = \boldsymbol{\Phi} \big[(1+c(\Delta t)) \boldsymbol{I} + \epsilon (\Delta t) \boldsymbol{\Lambda}\big]^{-1} \boldsymbol{\Phi}^\top.
	\end{equation}
	Then, writing $\boldsymbol{U}^{l+1}\in\mathbb{R}^{n\times m}$ in~\eqref{equ:U_lp1} with respect to the basis $\boldsymbol{\Phi}\in\mathbb{R}^{n\times n}$, $\boldsymbol{U}^{l+1} = \boldsymbol{\Phi} \boldsymbol{\tilde{V}}^{l+1}$ with the coefficient matrix $\boldsymbol{\tilde{V}}^{l+1}\in\mathbb{R}^{n\times m}$, results in
	\begin{equation*}
	\boldsymbol{\tilde{V}}^{l+1} = \big[ (1+c(\Delta t)) \boldsymbol{I} + \epsilon (\Delta t) \boldsymbol{\Lambda} \big]^{-1} \boldsymbol{\Phi}^\top \left( (1+c(\Delta t))\boldsymbol{U}^{l} - \frac{\Delta t}{2 \epsilon} \boldsymbol{\mathcal{T}}^{l} - (\Delta t) \, \boldsymbol{\omega} \, (\boldsymbol{U}^{l} - \boldsymbol{F}) \right).
	\end{equation*}
	Next, the eigendecomposition $\boldsymbol{\Lambda}\in\mathbb{R}^{n\times n}$, $\boldsymbol{\Phi}\in\mathbb{R}^{n\times n}$ of $\boldsymbol{L}_{\mathrm{sym}}$ is approximated by a truncated version $\boldsymbol{\Lambda}_k\in\mathbb{R}^{k\times k}$,  $\boldsymbol{\Phi}_k\in\mathbb{R}^{n\times k}$ using only the $k$ smallest eigenvalues of $\boldsymbol{L}_{\mathrm{sym}}$ (cf. \Cref{sec:binary_allen_cahn} for the binary case), and $\boldsymbol{B}^{-1}$ is projected into the corresponding eigenspace. This yields the modified iteration scheme
	\begin{equation}\label{equ:V_lp1}
	\boldsymbol{V}^{l+1} = \underbrace{\big[ (1+c(\Delta t)) \boldsymbol{I} + \epsilon (\Delta t) \boldsymbol{\Lambda}_k \big]^{-1} \boldsymbol{\Phi}_k^\top}_{=:\boldsymbol{Z}} \left( (1+c(\Delta t))\boldsymbol{U}^{l} - \frac{\Delta t}{2 \epsilon} \boldsymbol{\mathcal{T}}^{l} - (\Delta t) \, \boldsymbol{\omega} \, (\boldsymbol{U}^{l} - \boldsymbol{F}) \right).
	\end{equation}
	with the coefficient matrix $\boldsymbol{V}^{l+1}\in\mathbb{R}^{k\times m}$ and the new iterate $\boldsymbol{\tilde{U}}^{l+1} := \boldsymbol{\Phi}_k \boldsymbol{V}^{l+1}$.
	As the matrix $\big[ (1+c(\Delta t)) \boldsymbol{I} + \epsilon (\Delta t) \boldsymbol{\Lambda}_k \big]\in\mathbb{R}^{k\times k}$ is diagonal (with non-zero entries), its inverse and consequently the matrix $\boldsymbol{Z}\in\mathbb{R}^{k\times n}$ are easy to compute.
	Note that $\boldsymbol{Z}$ does not depend on the matrices~$\boldsymbol{U}^{l}$ and~$\boldsymbol{\mathcal{T}}^{l}$ from the previous iterations, which means that it can be precomputed before starting the first iteration $l=0$. 
	Similar to the binary case, however, the computational costs for matrix-matrix multiplications like $\boldsymbol{Z} \boldsymbol{\mathcal{T}}^{l}$ remain dominant, for instance involving the matrices $\boldsymbol{\Phi}_k^\top\in\mathbb{R}^{k\times n}$ and $\boldsymbol{\mathcal{T}}^{l}\in\mathbb{R}^{n\times m}$. 
\end{sloppypar}

Since we cannot expect that after an iteration step a row of $\boldsymbol{\tilde{U}}^{l+1} := \boldsymbol{\Phi}_k\boldsymbol{V}^{l+1}\in\mathbb{R}^{n\times m}$ is still an element of the Gibbs simplex~$\Sigma^{m}$, as defined in~\eqref{gibbssimplex}, we need to project the result of each iteration back to $\Sigma^{m}$. Omitting this projection can lead to a blow-up of the solution to iterates $U$ taking values far beyond the Gibbs simplex~$\Sigma^{m}$.
To this end, the technique proposed in \cite{chen2011projection} is employed. Here, for a given $\boldsymbol{\tilde{u}}_{i}^{\ell+1}\in\mathbb{R}^m$, we search for the element $\boldsymbol{\sigma}_i \in \Sigma^{m}$ with the minimal distance, i.e.\
\begin{equation*}
\boldsymbol{\sigma}_i = \text{arg} \min_{\boldsymbol{\sigma} \in \Sigma^{m}} \| \boldsymbol{\sigma} - \boldsymbol{\tilde{u}}_{i}^{\ell+1} \|,
\end{equation*}
and set $\boldsymbol{u}_{i}^{\ell+1}$ to $\boldsymbol{\sigma}_i$ afterwards.
Due to this projection to the Gibbs simplex $\Sigma^m$, we can interpret the components of~$\boldsymbol{u}_{i}$ as the empirical probabilities that the node $x_i\in\mathcal{V}$ belongs to each of the $m$ classes, or in other words, scores for class affiliation.

\section{Data generated by the stochastic block model}\label{sec:numerics:SBM}

We start by considering two example data sets generated by the multilayer stochastic block model (SBM) \cite{holland1983stochastic} which creates weight matrices with a prescribed clustering structure following a random distribution approach. The value $1$ in the binary weight matrix represents the presence of an edge between two nodes while $0$ means no edge. The parameters $p_{\mathrm{in}}, p_{\mathrm{out}} \in [0,1]$ represent the probabilities for the generation of an edge between two nodes belonging to the same class and to different classes, respectively. As $n$ is chosen to be small throughout this subsection, all matrices are computed and stored explicitly.

The first example illustrates that negative powers $p$ in the power mean Laplacian are robust against noisy data in some layers when other layers are informative while the power mean Laplacian with positive powers $p$ is not. The second example demonstrates that the power mean Laplacian combining the information of all layers is required for good classification results when all layers are informative. Furthermore, it compares the classification performance for different powers $p$ in the power mean Laplacian. Although we choose the same number of nodes per cluster throughout this section, a (moderately) different number of nodes per cluster does not affect the qualitative behavior.

In the first example we generate a multilayer graph with $T=2$ layers and $m=2$ classes each consisting of $n_{\mathrm{cluster}}=50$ nodes with both layers separating the two classes by the choice of the probabilities $p_{\mathrm{in}}=0.7$ and $p_{\mathrm{out}}=0.3$ in a first step. We employ \Cref{alg::multilayer_graph_allen_cahn_multiclass} using $k=2$ eigenpairs and $4\%$ pre-labeled points and compare its performance using $p=10$ and $p=-20$ averaging over $100$ random graphs. The Allen--Cahn parameters are set to $\epsilon=0.005$, $\omega_{0}=1\,000$, $c=\omega_{0}+3/\epsilon$, $\Delta t=0.01$, $\verb|max_iter|=300$ and $\verb|tolerance|=10^{-6}$. The classification errors of $0.2\%$ for $p=10$ and $0.03\%$ for $p=-20$ are both good. In a second step we keep the first layer informative while making the second layer noisy by assigning the equal probabilities $p_{\mathrm{in}}=p_{\mathrm{out}}=0.5$. The same Allen--Cahn classifier now produces a classification error of $46.1\%$ for $p=10$ while the negative power $p=-20$ performs much better with an error of only $1.8\%$. For comparison, $p=1$ yields an accuracy of $7.7\%$ in this example. Moreover, the choice of $p=-30$ produces an increased error of $4.3\%$ compared to the $p=-20$ case. These results confirm the observations made for spectral clustering in \cite{pmlr-v84-mercado18a} for the Allen--Cahn scheme, namely that negative powers in the power mean Laplacian tend to outperform positive powers and are robust against uninformative layers. In that light, the development of efficient numerical methods for the computation of the eigeninformation of the power mean Laplacian for the more difficult case $p>0$ appears unattractive.

\begin{figure}[ht]
	\begin{scriptsize}
		\begin{center}
			\begin{tabular}{|c|c c c | c c c |c|}
				\hline
				$p$\rule[-0.6em]{0em}{1.9em} &
				$\boldsymbol{L}_{\mathrm{sym}}^{(1)}$ & $\boldsymbol{L}_{\mathrm{sym}}^{(2)}$ & $\boldsymbol{L}_{\mathrm{sym}}^{(3)}$ & $\boldsymbol{L}_{p,\delta}^{(12)}$ & $\boldsymbol{L}_{p,\delta}^{(13)}$ & $\boldsymbol{L}_{p,\delta}^{(23)}$ & $\boldsymbol{L}_{p,\delta}$\\ \hline
				$10$\rule[0em]{0em}{1em} & 33.9 & 34.2 & 34.1 & 39.8 & 40.0 & 40.9 & 46.91\\
				$5$ & 33.9 & 34.2 & 34.1 & 33.8 & 34.2 & 34.3 & 10.86\\
				$1$ & 33.9 & 34.2 & 34.1 & 28.4 & 29.0 & 28.5 & 0.71\\
				\hline
				$-1$\rule[0em]{0em}{1em} & 33.9 & 34.2 & 34.1 & 22.3 & 22.6 & 23.6 & 0.33\\
				$-5$ & 33.9 & 34.2 & 34.1 & 15.9 & 16.0 & 17.0 & 0.17\\
				$-10$ & 33.9 & 34.2 & 34.1 & 9.8 & 10.1 & 9.5 & 0.09\\
				$-20$ & 33.9 & 34.2 & 34.1 & 3.6 & 3.7 & 3.5 & 0.02\\
				$-30$ & 33.9 & 34.2 & 34.1 & 4.4 & 4.1 & 4.0 & 0.47\\
				$-50$ & 33.9 & 34.2 & 34.1 & 27.9 & 26.8 & 26.1 & 13.25\\
				\hline
			\end{tabular}
			\hfill
			\begin{tikzpicture}[baseline={( 
				current bounding box.center)}]
			\begin{axis}[
			font=\footnotesize,
			enlarge x limits=true,
			height=0.26\textwidth,
			grid=major,
			xmajorgrids=false,
			width=0.42\textwidth,
			xtick={-30,-20,-10,-5,1,5,10},
			ytick={0.01,0.1,1,10,100},
			yticklabel={\pgfkeys{/pgf/fpu=true}\pgfmathparse{exp(\tick)}\pgfmathprintnumber[1000 sep={\,},fixed relative, precision=3]{\pgfmathresult}\pgfkeys{/pgf/fpu=false}},
			xmin=-30,xmax=10,
			ymin=0.01,ymax=100,
			ymode=log,
			xlabel={$p$},
			ylabel={clustering error in \%},
			legend style={legend cell align=left, at={(1.0,1.4)}}, 
			legend columns = -1,
			]
			
			\addplot[black,mark=triangle,mark size=2,mark options={solid}] coordinates {
				(10,39.8) (5,33.8) (1,28.4) (-1,22.3) (-5,15.9) (-10,9.8) (-20,3.6) (-30,4.4)
			};
			\addlegendentry{$\boldsymbol{L}_{p,\delta}^{(12)}$}
			\addplot[black,mark=o,mark size=2,mark options={solid,rotate=180}] coordinates {
				(10,40.0) (5,34.2) (1,29.0) (-1,22.6) (-5,16.0) (-10,10.1) (-20,3.7) (-30,4.1)
			};
			\addlegendentry{$\boldsymbol{L}_{p,\delta}^{(13)}$}
			\addplot[black,mark=square,mark size=2,mark options={solid,rotate=180}] coordinates {
				(10,40.9) (5,34.3) (1,28.5) (-1,23.6) (-5,17.0) (-10,9.5) (-20,3.5) (-30,4.0)
			};
			\addlegendentry{$\boldsymbol{L}_{p,\delta}^{(23)}$}
			\addplot[red,mark=x,mark size=2,mark options={solid}] coordinates {
				(10,46.91) (5,10.86) (1,0.71) (-1,0.33) (-5,0.17) (-10,0.09) (-20,0.02) (-30,0.47)
			};
			\addlegendentry{$\boldsymbol{L}_{p,\delta}$}
			\end{axis}
			\end{tikzpicture}
		\end{center}
	\end{scriptsize}
	\caption{Clustering errors in percent for different graph Laplacians on the 3 class SBM data set with $p_{\mathrm{in}}=0.7$ and $p_{\mathrm{out}}=0.3$, choosing $\delta=0$ for $p>0$ and $\delta=\log(1+|p|)$ for $p<0$. }\label{tab:SBM_3layer}
\end{figure}
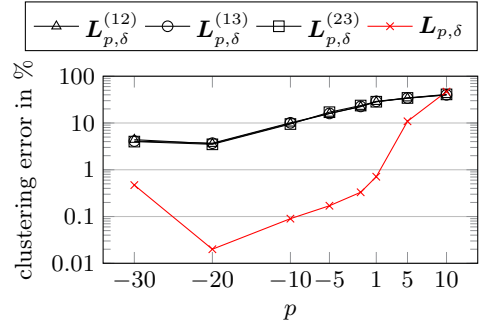

In the second example we consider a multilayer graph with $T=3$ layers and $m=3$ classes each consisting of $n_{\mathrm{cluster}}=50$ nodes. We choose $p_{\mathrm{in}}=0.7$ and $p_{\mathrm{out}}=0.3$ such that each layer $i=1,2,3$ separates the nodes belonging to cluster $i$ from the remaining two classes which distributes the necessary information for perfect graph segmentation across all three layers. We again apply the Allen--Cahn scheme and set all parameters as in the previous example. As mentioned in \Cref{sec:graph_allen_cahn_multilayer}, we predict the class taking the row-wise maximum of the output matrix $\boldsymbol{U}^l$. 
We pre-label $4\%$ of the nodes per class and compare the performance of the three single layer graph Laplacians $\boldsymbol{L}_{\mathrm{sym}}^{(1)}$, $\boldsymbol{L}_{\mathrm{sym}}^{(2)}$, $\boldsymbol{L}_{\mathrm{sym}}^{(3)}$, the three combinations of power mean Laplacians using two out of the three layers $\boldsymbol{L}_{p,\delta}^{(12)}$, $\boldsymbol{L}_{p,\delta}^{(13)}$, $\boldsymbol{L}_{p,\delta}^{(23)}$, and the power mean Laplacian using all three layers $\boldsymbol{L}_{p,\delta}$ in the multiclass Allen--Cahn scheme. In order to include an equal number of independent random graph layers in each case, we average over $300$ single layer graphs for $\boldsymbol{L}_{\mathrm{sym}}^{(1)}$, $\boldsymbol{L}_{\mathrm{sym}}^{(2)}$, $\boldsymbol{L}_{\mathrm{sym}}^{(3)}$, $150$ two layer graphs for $\boldsymbol{L}_{p,\delta}^{(12)}$, $\boldsymbol{L}_{p,\delta}^{(13)}$, $\boldsymbol{L}_{p,\delta}^{(23)}$ and $100$ three layer graphs for $\boldsymbol{L}_{p,\delta}$. For each graph Laplacian, we choose $k=3$ eigenpairs. The construction of the (single and multilayer) graph and the application of \Cref{alg::multilayer_graph_allen_cahn_multiclass} require approximately 0.03 seconds for each instance. We observed no significant difference in runtimes for the different graph Laplacians in this case where $n$ is relatively small. We visualize the average clustering errors, i.e., the relative differences between our predicted classes and the ground truth averaged over all random graphs, for different $p$ in \Cref{tab:SBM_3layer}.

The results illustrate that for $p \leq 1$ the multiclass Allen--Cahn scheme obtains very good classification results only when the information of all three layers is combined in the power mean Laplacian~$\boldsymbol{L}_{p,\delta}$. However, as $p>1$ increases, the power mean Laplacian's classification error increases considerably until it even surpasses the single layer errors. Interestingly, although it is not required for the invertibility of~$\boldsymbol{L}_{p,\delta}$, the inclusion of the diagonal shift $\delta>0$ in the case $p>1$ leads to distinctly improved classification results compared to using no shift (in any case, $p\leq 1$ still gives better results). For example, $\boldsymbol{L}_{p,\delta}$ with $p=10$ and $\delta=\log(1+|p|)$ only yields an error of $3.99\%$. For $p=1$ there is no difference in the classification result. As a shift and invert strategy applied to the power mean Laplacian, cf.\ \cite[Sec.~7.6.1]{golub2012matrix}, will typically be needed for the computation of the smallest eigenpairs of~$\boldsymbol{L}_{p,\delta}$, we conjecture that the incorporation of a positive shift already in the single layer Laplacians $\boldsymbol{L}_{\mathrm{sym},\delta}^{(t)}$ provides for more numerical stability in the eigeninformation computations. A more detailed investigation of this behavior is left to future research.

In the case $p \leq 5$, removal of informative layers leads to a significant loss in classification accuracy. While the informativity of different layers w.r.t.\ the clustering structure will generally not be equally distributed across different layers in real world data sets, this example still illustrates the advantage of the power mean Laplacian for multilayer graphs over classical single layer graph tools like the single layer graph Laplacian for suitable choices of $p$. For example, \Cref{tab:SBM_3layer} shows a monotonous decrease in the classification error of the power mean Laplacians for a descreasing power $p$ up until $p=-20$ where numerical effects start to perturb the results.

Finally, we remark that the further investigation of the improved classification accuracy of the shifted power mean Laplacian for positive $p$'s as well as the influence of adversarial noise instead of random noise in some graph layers in the setting of our first example in this section are interesting roads for future research.

\end{appendix}

\end{document}